%% file: acl_latex.tex
\documentclass[11pt]{article}

\usepackage[preprint]{acl}  %

\usepackage{times}
\usepackage{latexsym}

\usepackage[T1]{fontenc}
\usepackage[utf8]{inputenc}

\usepackage{microtype}
\usepackage{inconsolata}

\usepackage{amsmath}
\usepackage{amssymb}
\usepackage{amsfonts}
\usepackage{nicefrac}
\usepackage{booktabs}
\usepackage{multirow}
\usepackage{array}
\usepackage{graphicx}
\usepackage{subcaption}
\usepackage{enumitem}
\usepackage{xcolor}
\usepackage{xspace}
\usepackage{listings}
\lstdefinestyle{prompt}{
  basicstyle=\footnotesize\ttfamily,
  breaklines=true,
  breakindent=0pt,
  columns=fullflexible,
  keepspaces=true,
  frame=single,
  framesep=5pt,
  xleftmargin=4pt,
  xrightmargin=4pt,
}

\newcommand{\coarse}{`coarse\xspace}

\newcommand{\ci}[2]{\shortstack{#1\\[1.5pt]{\scriptsize [#2]}}}
\newcommand{\cnt}[2]{\shortstack{#1\\[1.5pt]{\scriptsize (#2)}}}

\title{Benchmarking Agentic Review Systems}

\author{
  Dang Nguyen$^{1}$\thanks{\, Corresponding author: \texttt{dangnguyen@uchicago.edu}} \quad Wanqing Hao$^{1}$ \quad Yanai Elazar$^{2}$ \quad Chenhao Tan$^{1}$ \\
  $^{1}$University of Chicago \quad $^{2}$Bar-Ilan University \\
}

\begin{document}
\maketitle

\begin{abstract}
A new class of agentic review systems are emerging as a remedy to the pressure placed on peer review systems by AI-assisted research, but it is unclear how they should be evaluated.
We evaluate two open-source systems (OpenAIReview and \coarse), one proprietary system (Reviewer3), and a zero-shot baseline, across six LLMs spanning frontier and efficient models.
First, we study whether AI reviews on ICLR/NeurIPS papers track with papers' quality as approximated by external signals such as citations and acceptance decisions.
Every system performs above chance in pairwise accuracy, and the best is OpenAIReview\,+\,GPT-5.5 at $83.0\%$.
Second, to test whether systems can catch 
errors with known ground truth, we construct a perturbation benchmark that injects four categories of errors
into papers across eight arXiv subject classes and measure detection recall.
The strongest configuration (OpenAIReview\,+\,GPT-5.5) catches $71.6\%$ of injected errors, leaving substantial room for improvement.
The union of detections across six models reaches $83.3\%$ recall,
suggesting different models detect different errors and better harness design can potentially increase performance.
Beyond these benchmarks, we study a public deployment of OpenAIReview with real users.
Votes on its comments skew positive at $1.44$ to $1$, and the most common complaints are about false positives and minor nitpicks.
Together, by evaluating full review systems backed by state-of-the-art models on real research papers, we show that while AI reviews still have room for improvement, they can already track human quality judgments well, catch important errors, and earn positive feedback from real users.
\end{abstract}

\input{intro}

\input{related}

\input{openaireview}
\input{background}

\input{conference}

\input{perturbation}

\input{analysis}

\input{user_feedback}
\input{discussion}

\bibliography{custom}

\appendix
\input{appendix}

\end{document}

%% file: intro.tex
\section{Introduction}

\begin{figure*}[!t]
  \centering
  \includegraphics[width=0.9\linewidth]{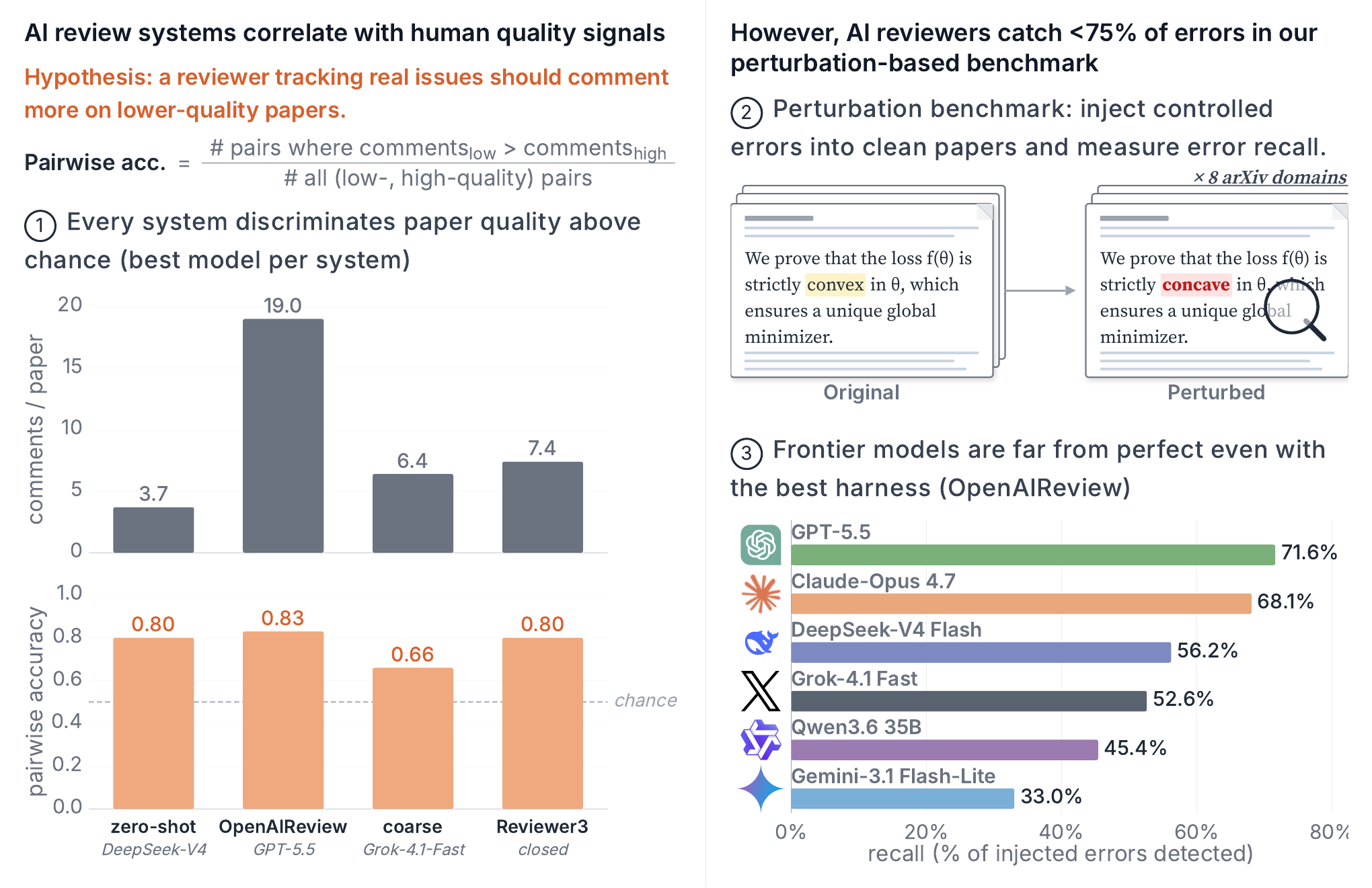}
  \caption{
    AI reviewer systems can generate useful reviews.
    \textbf{(1)} On ICLR/NeurIPS papers, the best backend for each system produces different comment volumes (top), yet every system discriminates paper quality above chance (bottom) at separating low- from high-quality papers.
    \textbf{(2)} We inject controlled errors into papers and check whether the reviewer flags the perturbed span.
    \textbf{(3)} The strongest configuration (OpenAIReview\,+\,GPT-5.5) catches $71.6\%$ of injected errors.
  }
  \label{fig:headline}
\end{figure*}

Peer review is under increasing pressure as AI-assisted papers increase submission volumes and flood the system~\citep{liu2026deathspiral,lu2024aiscientist}.
\citet{liu2026deathspiral} formalize this dynamic as a \emph{review death spiral}: as submissions overwhelm reviewer capacity, review accuracy degrades, acceptance becomes more random, and weaker papers are drawn in to ``try their luck,'' further overloading reviewers and pushing the system toward a sharp collapse.
Their analysis identifies that only improvements in \emph{review precision} (the ability to discriminate high from low-quality papers) can stabilize the system, making AI-assisted reviewing a necessary remedy.
Indeed, major conferences have begun integrating LLMs into the reviewing pipeline accordingly and found promising results~\citep{icml2026pat,biswas2026aaai}.

At the same time, LLM-based automated reviewer systems have emerged, with proprietary ones like Refine~\citep{refine} and Reviewer3~\citep{reviewer3}, and open-source ones like OpenAIReview~\citep{openaireview} and \coarse~\citep{coarse}.
These systems support multi-agent setups, structured prompts, and different forms of context management to emit detailed feedback on paper snippets rather than a score or acceptance decision.
Although there are encouraging anecdotes about their helpfulness, it is unclear how reviewer systems compare to each other.
Prior benchmarks have either been small in scale or evaluated raw LLMs rather than the system as a whole~\citep{liu2023reviewergpt,tyser2024aidriven,xi2025flaws}, and models have improved substantially ever since.
This marks a timely moment to revisit benchmarking AI reviewer systems with \textit{stronger models}, \textit{system-level comparisons}, and \textit{real papers}.

We first study whether AI review systems such as OpenAIReview, \coarse, and Reviewer3 can correlate with
paper quality from different sources, including community citations, conference decisions, reviewer scores, and a composite of the three.
For instance, high citation counts can signal a strong paper while low counts can signal a weaker paper.
We find that AI review systems register this signal on ICLR/NeurIPS papers despite not being explicitly trained to approximate acceptance decisions.
Under the assumption that weaker papers (according to the proxies)
should incur more comments, we compute the accuracy of systems on randomly sampled (low, high) quality paper pairs.
Every system assigns more comments to the low-quality paper above chance, and the signal is strongest on frontier models (OpenAIReview\,+\,GPT-5.5 reaches $0.83$ pairwise accuracy).
This trend also holds across models and increases with model strength, providing evidence that today's models can provide useful signal in reviewing.

However, models are far from perfect 
when it comes to identifying all errors in a paper.
Whereas the quality-proxy study above is confined to ICLR/NeurIPS submissions, here we move beyond a single field and construct a comprehensive \emph{perturbation benchmark}, where we introduce errors into otherwise-good papers and evaluate AI review systems' recall of these errors.
The benchmark spans eight arXiv subject classes ranging from Econometrics to Genomics, with four error categories: local math
edits, false claims, faulty reasoning, and experimental design or analysis errors.
The strongest configuration (OpenAIReview backed by GPT-5.5) catches $71.6\%$ of injected errors (Figure~\ref{fig:headline}, right).
OpenAIReview's largest gains over a zero-shot baseline are on prose-level errors (e.g., faulty reasoning increases from 36.8\%$\to$68.4\%), while math-token edits show smaller gains.
This is in-line with OpenAIReview's running summarization design to support reviewing longer papers, since prose-level errors can often span a longer context than local math edits.

Despite the not-yet-perfect recall, gains are possible with better system design.
Different backend models under OpenAIReview are complementary: the recall from all models combined reaches $83.3\%$, $11.7$ points above GPT-5.5. 
Together, these findings point at the potential for AI review systems to be adopted in conference reviewing and for new systems to be designed that can do well on all evaluation dimensions.

We then move from controlled benchmarks to real use.
We deploy OpenAIReview as a public web tool and collect feedback on $1{,}360$ reviews of $1{,}100$ papers.
Users vote positively on the comments, with likes outnumbering dislikes $1.44$ to $1$, and many mark comments as addressed, evidence that the reviews carry value in practice.
Sorting the downvoted comments by reason shows the main weakness is comment precision: most complaints are about false positives and minor nitpicks.

To summarize, our contributions are as follows:
\begin{itemize}%
  \item We introduce a perturbation-based benchmark and find that the strongest system catches about $71.6\%$ of injected errors. 
  \item We show that today's AI reviewer systems pick up on signals of paper quality at $66-83\%$ accuracy without being explicitly trained to, with a stronger signal in stronger models.
  \item We find that models often comment on disjoint sets of paragraphs and combining them can result in higher recall.
  \item In a public deployment of OpenAIReview, users vote positively on its comments ($1.44$ to $1$ likes to dislikes), and the main weakness is comment precision, with most complaints being false positives and minor nitpicks.
\end{itemize}

%% file: related.tex
\section{Related Work}
\label{sec:related}

\paragraph{Automated review systems.}
Earlier work targeted LLM-assisted reviewing with smaller backbone models, ranging from aspect-level summarization~\citep{yuan2022reviewing} to direct GPT-4 feedback on full papers~\citep{liang2023feedback}.
Recent systems wrap frontier LLMs in multi-stage harnesses (section-level multi-agent pipelines~\citep{darcy2024marg}, rubric-scored sequential prompts~\citep{tyser2024aidriven}, structured-reasoning agents trained on a dedicated review chain-of-thought corpus~\citep{gao2025reviewagents}, and full research pipelines that embed reviewing as a downstream stage~\citep{lu2024aiscientist}) to emit detailed advisory feedback rather than accept/reject decisions.
Our benchmark targets a different slice of this space: rather than the academic prototypes cited above, we evaluate the publicly available reviewer systems that authors can actually run today (OpenAIReview~\citep{openaireview}, \coarse~\citep{coarse}, and the commercial Reviewer3~\citep{reviewer3}), together with a zero-shot single-prompt baseline, and we score each as a whole pipeline rather than swapping in its underlying LLM.

\paragraph{Perturbation-based error detection.}
Injecting controlled errors and measuring whether reviewers catch them is a recurring evaluation idea with a long NLP lineage~\citep{gardner2020contrast,kaushik2020learning,olmpics,ribeiro2020checklist,kassner2020negated,sai2021perturbation}.
For paper review specifically, early work hand-injected small numbers of errors into short papers~\citep{liu2023reviewergpt,tyser2024aidriven}, and concurrent benchmarks FLAWS~\citep{xi2025flaws} and SPECS~\citep{biswas2026aaai} scale this up to ICML and AAAI papers.
We differ in perturbing by error \emph{type} (math, claim, reasoning, experimental) rather than review aspect, varying the model across six LLMs, and benchmarking independently-developed third-party systems head-to-head rather than raw LLM prompts.

\paragraph{LLMs vs.\ humans.}
Several recent works compare LLM and human reviews: \citet{liang2023feedback} find GPT-4 captures consensus critiques but emphasizes broader implications while underweighting novelty. \citet{li2025unveiling} report LLM reviews skew toward strengths over weaknesses and barely scale critique complexity with paper quality. \citet{gao2025reviewagents} introduce ReviewBench for direct LLM-vs.-human comparison.
Our quality-proxy correlation study contributes a complementary signal, namely comment behavior on real ICLR/NeurIPS papers across four quality proxies, three reviewer systems, and six models.

%% file: openaireview.tex
\section{OpenAIReview}
\label{sec:openaireview}

OpenAIReview~\citep{openaireview} is an open-source system that takes a full paper as input and returns a list of comments, each connected to a quoted passage.
We benchmark it against existing systems in Sections~\ref{sec:conference} and~\ref{sec:perturbation} and analyze real user feedback from a deployed version in Section~\ref{sec:production}.

\begin{figure*}[t]
  \centering
  \includegraphics[width=0.9\textwidth]{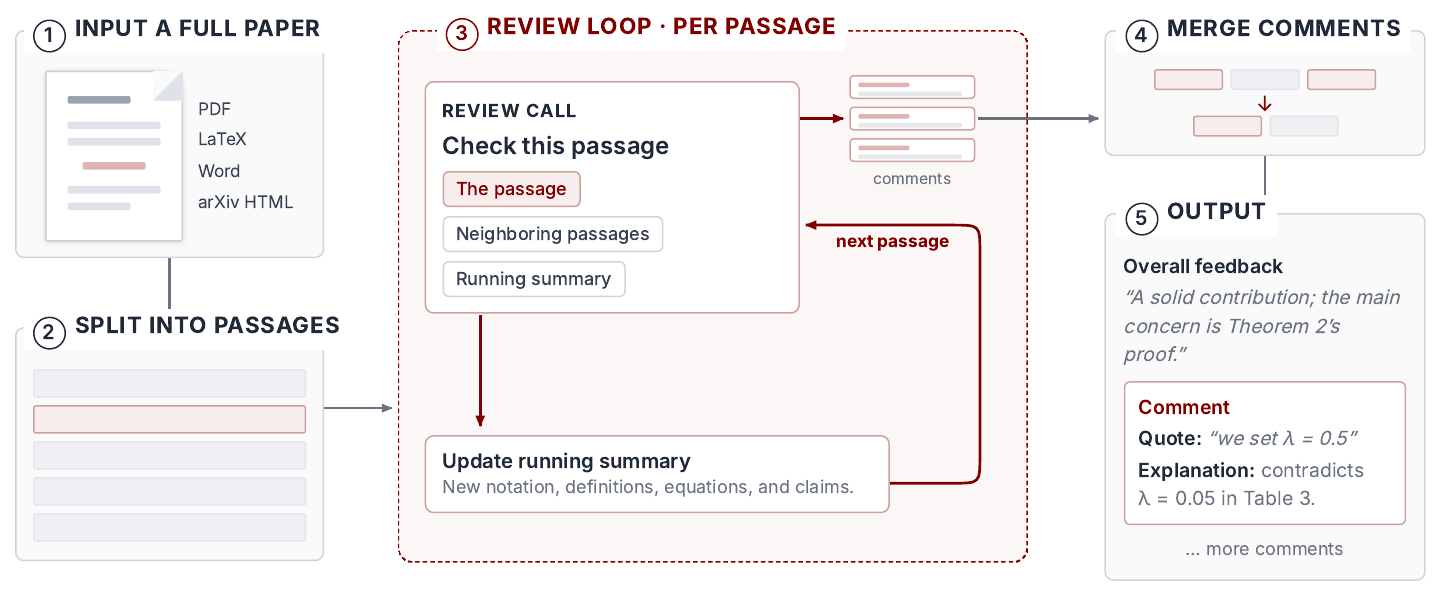}
  \caption{OpenAIReview reviews a paper one passage at a time, checking each passage against its neighbors and a running summary of the paper so far. The summary is updated after each passage, and a final call merges the collected comments.}
  \label{fig:openaireview-flowchart}
\end{figure*}

\paragraph{Paper processing pipeline.}
The system accepts papers 
in different formats, such as PDF, markdown, or LaTeX.
The review begins with an overall feedback section, followed by a list of comments, each containing a quoted passage and an explanation of the issue (Figure~\ref{fig:openaireview-example} in Appendix~\ref{app:openaireview-impl} shows an example and Figure~\ref{fig:ui-openaireview} shows the web interface). 
Figure~\ref{fig:openaireview-flowchart} illustrates how the system produces a review: the paper is first split into passages of roughly equal length, and the system reviews one passage at a time.
Each passage is checked against two sources of context: a window of neighboring passages and a running summary of everything read so far.
The running summary accumulates notation, definitions, key equations, theorems, assumptions, and claims. 
After each passage is checked, a separate model call updates the summary with any new content from it.
This lets the model catch issues that span long distances, such as a symbol used inconsistently with a definition given several sections earlier, without placing the entire paper into a single prompt.
Finally, 
the system drops duplicates and merges comments that point to the same underlying issue.
We use the same backbone model across the pipeline.

\paragraph{Review prompts.}
The main review prompt gives the model a fixed list of checks to run on each passage, e.g., mathematical and formula errors, inconsistent notations, 
overstated claims, 
and methods described too vaguely to reproduce. 
To reduce false positives,
the prompt tells the model to first check whether a concern is resolved by the surrounding context, and to skip categories such as formatting issues or forward references.
Figure~\ref{fig:review-prompt} in Appendix~\ref{app:openaireview-impl} shows the full review prompt, and Figures~\ref{fig:summary-prompt} and~\ref{fig:aux-prompts} give the prompts for the summary update, consolidation, and overall feedback.

\paragraph{Evaluating a review.}
As mentioned, a review has two parts: an overall feedback section and a list of individual comments.
This is the format that recent review systems such as Refine~\citep{refine} and \coarse~\citep{coarse} share, so how to evaluate it is a general question beyond just OpenAIReview.
The overall feedback gives a high-level assessment of the paper's quality, clarity, and main issues.
Judging whether it is accurate and helpful is a separate question, likely best left to an LLM judge, and we set it aside here.
We focus on the comments, which make concrete claims about particular passages.
A comment is only useful if it points to a real problem, so we evaluate comments based on two criteria: whether weaker papers draw more comments, and more serious ones (Section~\ref{sec:conference}), and whether the comments catch errors we know are present (Section~\ref{sec:perturbation}).

%% file: background.tex
\section{Systems Overview}
\label{sec:systems}

Before presenting our two evaluations, we describe the review systems we benchmark and the LLMs that power them. The same set of systems and models is used in both Section~\ref{sec:conference} and Section~\ref{sec:perturbation}.

\begin{figure*}[t]
  \centering
  \begin{subfigure}[b]{0.66\textwidth}
    \centering
    \includegraphics[width=\linewidth]{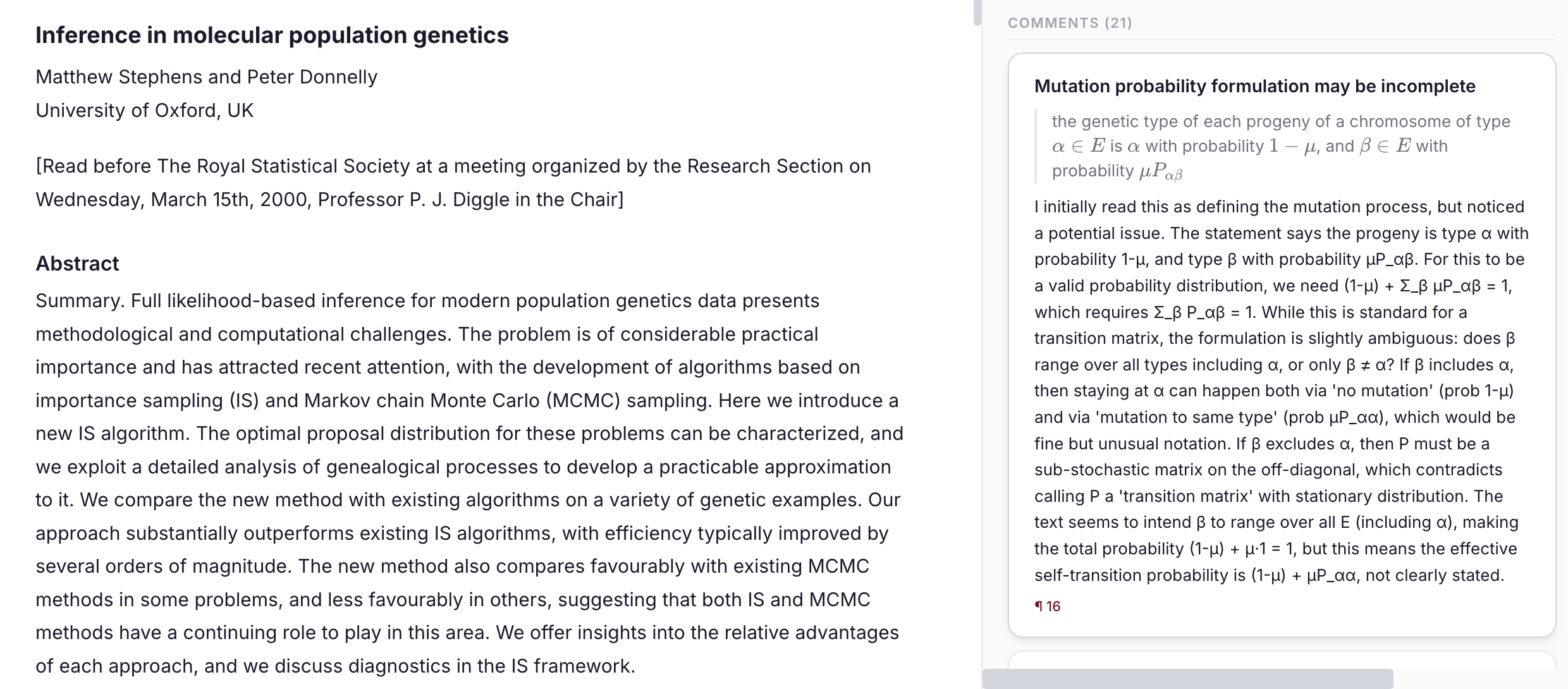}
    \caption{OpenAIReview}
    \label{fig:ui-openaireview}
  \end{subfigure}
  \hfill
  \begin{subfigure}[b]{0.30\textwidth}
    \centering
    \includegraphics[width=\linewidth]{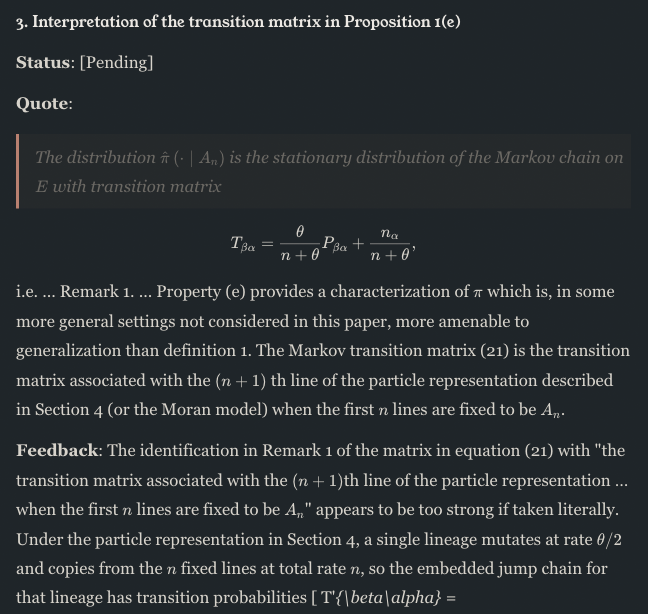}
    \caption{\coarse}
    \label{fig:ui-coarse}
  \end{subfigure}
  \caption{Review interfaces of the two automated systems we benchmark on the same paper. Both surface a list of comments, each anchored to a passage in the paper, that we use as the unit of analysis throughout this work.}
  \label{fig:ui-comparison}
\end{figure*}

\paragraph{Review methods.}
We compare four review systems, all of which take the full paper as input and emit a list of comments, where each comment contains a quoted passage and an explanation of the issue (two of the open systems' interfaces are shown in Figure~\ref{fig:ui-comparison}):
\begin{itemize}[nosep]
  \item \emph{Zero-shot.} A single prompt asks the model to review the entire paper in one pass and return all issues it finds.
  \item \emph{OpenAIReview}~\citep{openaireview}. A system that processes the paper sequentially, maintaining a running summary of definitions, equations, and key claims. For each passage it checks the current text against the running summary and surrounding context to flag inconsistencies. 
  \item \emph{\coarse}~\citep{coarse}. An open-source multi-agent paper reviewer that combines a macro-level overview agent with parallel per-section agents (and adversarial proof verification on math-heavy sections), followed by an editorial pass that deduplicates and filters the merged comment set.
  \item \emph{Reviewer3}~\citep{reviewer3}. A closed-source commercial reviewer system that emits a small set of high-priority comments per paper. Its internal model and prompts are not exposed, so it is treated as a single fixed system and not paired with the LLM sweep below.
\end{itemize}

\paragraph{Models.}
Zero-shot, OpenAIReview, and \coarse can each be backed by a different LLM.
We evaluate the three open systems across two frontier models: GPT-5.5~\citep{openai2026gpt55} and Claude Opus 4.7~\citep{anthropic2026claudeopus47}, and four efficient models: DeepSeek-V4-Flash~\citep{deepseek2026v4flash}, Qwen3.6-35B-A3B~\citep{qwen2026qwen36}, Gemini-3.1-Flash-Lite~\citep{google2026geminiflashlite}, and Grok-4.1-Fast~\citep{xai2026grok41fast}, all accessed via OpenRouter without reasoning mode.

\paragraph{Evaluation studies.}
With the methods and models fixed, we evaluate them along two complementary axes.
The first evaluation (Section~\ref{sec:conference}) feeds each (method, model) pair real ICLR/NeurIPS papers 
and tests whether the number of issues raised tracks paper quality as approximated by reception signals (citations, awards, review scores).
The perturbation benchmark (Section~\ref{sec:perturbation}) takes the same systems and tests whether they detect controlled errors injected into otherwise-clean papers.
Beyond ICLR/NeurIPS papers, it spans eight arXiv subject classes, from Econometrics to Genomics, in order to provide diverse ground-truth errors across different theoretical and empirical fields.

%% file: conference.tex
\section{AI Review Systems Correlate with Human Quality Signals}
\label{sec:conference}

\subsection{Method}

We compare automated review systems' outputs on real ICLR/NeurIPS papers to test whether they correlate with paper quality as judged by the research community through publication, citation, and review score signals.
Because paper quality has no gold-standard label, we instead use four \emph{quality proxies} at different evaluation scales derived from publication, citation, and review score signals.
Within each proxy we select 30 high-quality and 30 low-quality papers from ICLR/NeurIPS.
We filter for papers with substantive reviews, meaning at least three official reviews with a non-null average score.
The four quality proxies are defined at different scales:
\begin{itemize}
    \setlength{\itemsep}{1pt}
    \item \textbf{Community-level}: the top 30 papers by citations-per-year versus 30 randomly sampled rejected papers with no subsequent publication venue (construction details for all proxies in Appendix~\ref{app:conference-sampling}).
    \item \textbf{Conference-level}: 30 randomly accepted papers that were highlighted (Outstanding, Best, Oral, Spotlight) versus 30 randomly sampled rejected papers.
    \item \textbf{Reviewer-level}: among papers with substantive reviews, the top 30 by mean review score versus the bottom 30.
    \item \textbf{Composite}: the top 30 awarded papers ranked by the sum of citation rank and score rank, versus the bottom 30 papers (rejected, never-published) ranked by mean review score.
\end{itemize}
We emphasize that citations, awards, and review scores are noisy proxies of quality. 
We select the top and bottom groups as a tractable approximation, not as ground-truth measures of paper quality.

\paragraph{Paper selection.}
We draw papers from SNOR~\citep{snor}, a dataset that links OpenReview submissions for ICLR (2017--2025) and NeurIPS (2021--2025) to Semantic Scholar with citation counts, decisions, and reviewer scores.
We restrict to ICLR and NeurIPS 2021--2022, the earliest years for which both venues posted submissions on OpenReview, and far enough in the past that citation counts have had time to accumulate.
Given each low/high quality group has 30 papers, we get
60 papers per proxy and 240 in total (197 unique, since some papers satisfy more than one criterion).
For frontier models too expensive to run on the full set, we use a \emph{frontier subset} of 74 unique papers
and mainly report results on this subset.
Each paper is reviewed by every (method, model) pair from Section~\ref{sec:systems} on the first 10 pages.
Sampling design and full results are in Appendix~\ref{app:conference-full-cohort}.

\paragraph{Metrics.}
For each (method, model, quality proxy) triple, let $\bar{c}_{\rm high}$ and $\bar{c}_{\rm low}$ be the mean comments per paper on the high- and low-quality groups.
We report the \emph{pairwise accuracy}: the percentage of low- and high-quality pairs where the low-quality paper incurs more comments than the high-quality one, with ties (equal comment counts) counting as 0.5 (Appendix~\ref{app:auc-interpretation}).
We also report $\Delta = \bar{c}_{\rm low} - \bar{c}_{\rm high}$ and the percentage increase $\Delta / \bar{c}_{\rm high}$, expecting $\Delta > 0$ if the system tracks real issues.
Bracketed quantities are 95\% confidence intervals from a cluster bootstrap over papers (Appendix~\ref{app:ci-accuracy}).

\subsection{Results}

\begin{table}[t]
  \centering
  \resizebox{\columnwidth}{!}{%
  \begin{tabular}{lccccc}
    \toprule
    \bfseries Model & \bfseries $\bar{c}_{\rm high}$ & \bfseries $\bar{c}_{\rm low}$ & \bfseries $\Delta$ & \bfseries \% $\Delta$ & \bfseries Acc. \\
    \midrule
    GPT-5.5                       & 15.78 & 22.15 & $\mathbf{6.38}$ & $40.4\%$           & \ci{$\mathbf{0.83}$}{0.73, 0.91} \\
    Claude Opus 4.7               & 11.12 & 14.62 & $3.50$          & $31.5\%$           & \ci{$0.74$}{0.63, 0.84} \\
    Grok-4.1-Fast                 & 2.20  & 6.80  & $4.60$          & $\mathbf{209.1\%}$ & \ci{$0.80$}{0.70, 0.89} \\
    DeepSeek-V4-Flash             & 12.20 & 13.95 & $1.75$          & $14.3\%$           & \ci{$0.60$}{0.47, 0.72} \\
    Qwen3.6-35B-A3B               & 12.72 & 14.43 & $1.70$          & $13.4\%$           & \ci{$0.62$}{0.49, 0.74} \\
    Gemini-3.1-Flash-Lite         & 13.07 & 14.12 & $1.05$          & $8.0\%$            & \ci{$0.61$}{0.48, 0.74} \\
    \bottomrule
  \end{tabular}%
  }
  \caption{Under OpenAIReview, GPT-5.5 picks up on paper quality signal the best. $\bar{c}_{\rm high}$ and $\bar{c}_{\rm low}$ are mean comments per paper on the high- and low-quality groups (averaged across the four proxies). 
  }
  \label{tab:conference-model-aggregate}
\end{table}

\begin{table}[t]
  \centering
  \resizebox{\columnwidth}{!}{%
  \begin{tabular}{lccccccc}
    \toprule
    \bfseries Method & \bfseries $\bar{c}$ & \bfseries Comm. & \bfseries Conf. & \bfseries Rev. & \bfseries Comp. & \bfseries Overall \\
    \midrule
    \shortstack[l]{zero-shot    \\ {\itshape DeepSeek-V4}}   & $3.7$  & \ci{$\mathbf{0.93}$}{0.78, 1.00} & \ci{$0.60$}{0.34, 0.84} & \ci{$\mathbf{1.00}$}{1.00, 1.00} & \ci{$0.69$}{0.43, 0.92} & \ci{$0.80$}{0.71, 0.89} \\
    \shortstack[l]{OpenAIReview \\ {\itshape GPT-5.5}}       & $19.0$ & \ci{$0.87$}{0.69, 0.99} & \ci{$0.66$}{0.40, 0.89} & \ci{$0.94$}{0.80, 1.00} & \ci{$\mathbf{0.84}$}{0.64, 0.98} & \ci{$\mathbf{0.83}$}{0.73, 0.91} \\
    \shortstack[l]{\coarse      \\ {\itshape Grok-4.1-Fast}} & $6.4$  & \ci{$0.69$}{0.43, 0.90} & \ci{$0.60$}{0.33, 0.82} & \ci{$0.78$}{0.57, 0.94} & \ci{$0.59$}{0.33, 0.84} & \ci{$0.66$}{0.54, 0.78} \\
    \shortstack[l]{Reviewer3   \\ {\itshape (closed)}}      & $7.4$  & \ci{$0.78$}{0.50, 1.00} & \ci{$\mathbf{0.70}$}{0.45, 0.91} & \ci{$0.89$}{0.71, 1.00} & \ci{$0.81$}{0.60, 0.96} & \ci{$0.80$}{0.69, 0.89} \\
    \bottomrule
  \end{tabular}%
  }
  \caption{
    OpenAIReview\,+\,GPT-5.5 is the best model and system combination on overall pairwise accuracy.
    $\bar{c}$ is mean comments per paper. 
    \textbf{Overall} is the pooled accuracy across all four proxies.}
  \label{tab:conference-system-deltas}
\end{table}

\paragraph{GPT-5.5 picks up the quality signal best overall while Grok-4.1-Fast leads efficient models.}
Table~\ref{tab:conference-model-aggregate} ranks models under OpenAIReview on the 74-paper frontier subset.
GPT-5.5 leads at $0.83$ pairwise accuracy and $\Delta\,{=}\,{+}6.38$ comments, with Claude Opus 4.7 next at $0.74$.
Among efficient models, Grok-4.1-Fast tops the rest at $0.80$ on only $\bar{c}\approx 5$ comments per paper, with the other three models clustered between $0.60$ and $0.62$.
The same ranking holds on the full 240-paper set (Appendix~\ref{app:conference-full-cohort}).
Severity stratification confirms that weaker papers receive more \emph{severe} comments, not merely more: every model under OpenAIReview performs above chance on both major ($0.59$--$0.78$) and moderate ($0.61$--$0.91$) tiers, with GPT-5.5 again leading (Major $0.78$, Moderate $0.91$).
We omit the Minor tier, which correlates poorly with quality.
Per-system, per-tier breakdowns are in Appendix~\ref{app:conference-tier-breakdown}.

\paragraph{OpenAIReview\,+\,GPT-5.5 is the strongest configuration.}
Table~\ref{tab:conference-system-deltas} compares each system's best (method, model) pair, plus Reviewer3.
OpenAIReview\,+\,GPT-5.5 tops the overall accuracy at $0.83$, just ahead of zero-shot\,+\,DeepSeek-V4-Flash and Reviewer3 (both $0.80$).
\coarse trails at $0.66$ even with its best backend.
Zero-shot's accuracy comes on only $\bar{c}\approx 3.7$ comments per paper, which is directionally correct but likely too sparse to be useful in practice.
Per proxy, OpenAIReview and Reviewer3 are the most balanced (every proxy $\geq 0.66$ and $\geq 0.70$ respectively), while zero-shot peaks at the Reviewer proxy ($1.00$) but falls to $0.60$ on Conference.
Full breakdowns are in Appendix~\ref{app:conference-full-cohort}.

\paragraph{Takeaway.}
Comment volume tracks paper quality above chance across every system, model, and quality proxy, and the pattern holds when we restrict to major and moderate comments only.
This is despite none of these systems being trained or prompted to predict acceptance decisions: they are simply asked to surface issues.
Pairwise accuracy on comment counts is admittedly a coarse metric, and may miss cases where a concise review points out a single fatal issue.
Still, the consistency across systems, models, proxies, and severity tiers suggests today's models pick up real quality signals as a by-product of issue-finding, without task-specific finetuning.

%% file: perturbation.tex
\section{Perturbation Benchmark}
\label{sec:perturbation}

Quality proxies tell us whether comment volume tracks paper quality, but not whether individual comments are correct. Real papers lack such ground truth.
We complement the previous analysis with a controlled benchmark: inject known errors into clean papers and measure per-comment recall.

\subsection{Method}

We inject different types of errors into clean papers.
\textbf{Surface} errors are local math edits (sign flips, index/subscript changes, numeric edits, computation errors) that a reader can catch within a single equation.
The other three categories are \emph{prose-level} errors that require understanding context across a paragraph or paper: \textbf{Claim} (false theoretical or empirical statements), \textbf{Logic} (broken reasoning in proofs and arguments: circular reasoning, invalid implication, induction errors, missing cases), and \textbf{Experimental} (flawed experimental design: reversed causality, misinterpreted results, p-hacking).
Examples of each error are in Table~\ref{tab:error-taxonomy-full}, Appendix~\ref{app:benchmark-creation}.
Benchmark construction proceeds in five stages: \textit{extract}, \textit{generate}, \textit{validate}, \textit{verify}, \textit{inject}, described below.

\begin{table}[t]
  \centering
  \resizebox{\columnwidth}{!}{%
  \begin{tabular}{lccc}
    \toprule
    \bfseries Model & \bfseries zero-shot & \bfseries \coarse & \bfseries OpenAIReview \\
    \midrule
    GPT-5.5                       & \ci{59.8\%}{52.4, 67.3} & ---                     & \ci{\textbf{71.6\%}}{65.6, 76.9} \\
    Claude-Opus-4.7               & \ci{54.3\%}{49.0, 59.5} & ---                     & \ci{\textbf{68.1\%}}{62.8, 73.8} \\
    Grok-4.1-Fast                 & \ci{31.4\%}{26.9, 35.7} & \ci{12.8\%}{9.6, 16.0}  & \ci{\textbf{52.6\%}}{46.5, 59.1} \\
    DeepSeek-V4-Flash             & \ci{31.0\%}{26.2, 35.6} & \ci{20.7\%}{16.6, 25.0} & \ci{\textbf{55.8\%}}{50.0, 61.1} \\
    Qwen3.6-35B-A3B               & \ci{31.4\%}{26.6, 36.6} & \ci{18.2\%}{14.7, 21.8} & \ci{\textbf{45.4\%}}{39.9, 51.2} \\
    Gemini-3.1-Flash-Lite         & \ci{14.7\%}{11.3, 18.0} & \ci{12.3\%}{9.3, 15.1}  & \ci{\textbf{33.0\%}}{28.1, 37.2} \\
    \midrule
    \itshape Reviewer3 (closed)   & \multicolumn{3}{c}{\ci{26.5\%}{20.6, 32.9}} \\
    \bottomrule
  \end{tabular}%
  }
  \caption{OpenAIReview\,+\,GPT-5.5 attains the best recall out of all systems and models. Recall on the 24-paper frontier subset; brackets are 95\% bootstrap CIs over papers (Appendix~\ref{app:ci-recall}). Dashes indicate runs not executed for that model. Reviewer3 has no model selector and is reported as a separate row.
  }
  \label{tab:recall-overall}
\end{table}

\begin{table}[t]
  \centering
  \resizebox{\columnwidth}{!}{%
  \begin{tabular}{lccccc}
    \toprule
    \bfseries Method & \bfseries Overall & \bfseries Exper. & \bfseries Claim & \bfseries Reason. & \bfseries Surface \\
    \midrule
    \shortstack[l]{zero-shot    \\ {\itshape GPT-5.5}}     & \ci{59.8\%}{52.7, 67.2} & \ci{67.0\%}{50.0, 80.8} & \ci{67.0\%}{58.6, 74.5} & \ci{60.0\%}{57.1, 66.7} & \ci{45.8\%}{31.5, 60.3} \\
    \shortstack[l]{\coarse      \\ {\itshape DeepSeek-V4}} & \ci{20.7\%}{16.6, 24.9} & \ci{26.3\%}{21.3, 31.1} & \ci{19.6\%}{12.8, 26.6} & \ci{40.0\%}{35.7, 50.0} & \ci{16.0\%}{8.5, 24.2} \\
    \shortstack[l]{OpenAIReview \\ {\itshape GPT-5.5}}     & \ci{\textbf{71.6\%}}{65.9, 77.0} & \ci{\textbf{85.2\%}}{77.0, 92.0} & \ci{\textbf{83.3\%}}{75.2, 89.9} & \ci{\textbf{70.0\%}}{64.3, 83.3} & \ci{\textbf{47.3\%}}{33.8, 60.9} \\
    \shortstack[l]{Reviewer3    \\ {\itshape (closed)}}    & \ci{26.5\%}{20.2, 32.8} & \ci{35.4\%}{25.7, 45.7} & \ci{28.1\%}{18.6, 38.3} & \ci{10.0\%}{0.0, 33.3} & \ci{18.8\%}{11.2, 27.7} \\
    \bottomrule
  \end{tabular}%
  }
  \caption{OpenAIReview wins on every error category. For each system we report its best-performing backend on the 24-paper frontier subset (where GPT-5.5 and Claude-Opus-4.7 were also run). Reviewer3 has no model selector. 
  Brackets are 95\% bootstrap CIs over papers (Appendix~\ref{app:ci-recall}.
  }
  \label{tab:recall-by-type}
\end{table}

\paragraph{Paper selection.}
We sample 74 papers from 8 arXiv subject classes spanning theoretical and empirical research: Computational Complexity, Machine Learning, Econometrics, Experimental High-Energy Physics, Mathematics, Atomic and Cluster Physics, Genomics, and Applied Statistics.
For each class, we collect 10 arXiv submissions with full \LaTeX{} source and discard those whose source fails to compile or lacks the structural cues (equations, theorems, claim/argument paragraphs) that the extractor needs.
Filtering yields 5--10 papers per class for a total of 74.
For frontier models too expensive to run on the full set, we use a 24-paper subset (3 per domain) and report main-text results on this subset.

\textbf{(1) Extract.} A deterministic extractor scans the \LaTeX{} source for candidate perturbation sites: equations, definitions, theorems, and proofs in theoretical papers, or equations, claim/argument paragraphs, and experimental paragraphs in empirical papers. Each is annotated with the error categories admissible for that span (Table~\ref{tab:error-span}).
\textbf{(2) Generate.} A generator LLM
picks a subset of candidates and emits the replacement \LaTeX{}
and an explanation why the perturbation results in an error.
An audit of this model-driven selection against the full candidate pool finds only modest bias relative to random selection, mainly a preference for longer equations (Appendix~\ref{app:selection-bias}).
\textbf{(3) Validate.} A structural validator drops perturbations that are identical to the original, overlap an accepted edit, or break \LaTeX{} at the span boundaries.
\textbf{(4) Verify.} A checklist verifier then filters out edits that are indistinguishable from typos (e.g., the value is not used downstream) or that do not constitute an error (e.g., changing a parameter in a way that still satisfies its bounds).
\textbf{(5) Inject.} Surviving edits are applied to yield a single corrupted paper.

To confirm that the kept perturbations are genuine, well-formed errors, we manually audited a stratified sample of $40$ injected perturbations ($10$ per error type, covering every subtype and all 8 domains).
We judge $33/40$ ($82.5\%$) to be valid errors, $2$ ($5\%$) to be not true errors, and $5$ ($12.5\%$) to be ambiguous, with the latter two categories concentrated in the surface-numeric and empirical-claim subtypes.
Models, prompts, automated and manual validation details are in Appendix~\ref{app:pipeline-details}.

\subsection{Evaluation}
\label{sec:perturb-eval}

For each ground-truth perturbation, we determine detection via a two-stage filter applied to the reviewer's emitted comments: first, a \emph{fuzzy substring match} requires the perturbed text to be approximately contained in a review comment's quote above a threshold $\tau$ (which we validated in Table~\ref{tab:threshold-sweep}). Second, an \emph{LLM judge} rates whether the reviewer's explanation identifies the same error as the ground-truth perturbation, and passing comments are counted as detections.
Recall is then computed as the fraction of injected perturbations detected. Threshold values, judge model, and rating cutoff are given in Appendix~\ref{app:scoring}.

\subsection{Results}

\paragraph{OpenAIReview outperforms both zero-shot and \coarse with different backend models.}
Table~\ref{tab:recall-overall} shows overall recall on the 24-paper frontier subset, with 95\% bootstrap CIs over papers (see CI calculation in Appendix~\ref{app:ci-recall}).
OpenAIReview wins across the board: the largest absolute gain over zero-shot comes from DeepSeek-V4-Flash (+24.8 points), and the smallest gap is for GPT-5.5 (+11.8 points), which already attains 59.8\% under zero-shot.
The same pattern holds on the full 74-paper benchmark, where the four efficient models ran on every paper (Table~\ref{tab:recall-overall-full} in Appendix~\ref{app:perturb-full-cohort}).
Notably, \coarse and Reviewer3 both underperform zero-shot: \coarse falls below zero-shot on every shared backend ($20.7\%$ vs.\ $31.0\%$ on DeepSeek-V4-Flash), and Reviewer3's $26.5\%$ sits below every zero-shot row except Gemini-3.1-Flash-Lite. 
This is likely because these two systems optimize for editorial prioritization rather than exhaustive enumeration.
Both systems emit 5--10 highest-priority comments per paper, so recall on injected errors might not capture their full value as reviewers.

\paragraph{Prose-level errors show the largest OpenAIReview gains.}
Table~\ref{tab:recall-by-type} breaks down recall by error category on the same setup.
OpenAIReview's gains over zero-shot are concentrated on prose-level errors (claim, reasoning, experimental, each $+10$--$18$ points), while surface math-token recall remains similar ($45.8\%\to47.3\%$), suggesting that the running summarization fails to help with errors that can already be detected locally.
\coarse is weaker than every other system on prose-level errors at $19.6\%$ on claim errors and $26.3\%$ on experimental errors.
While it does best on reasoning ($40.0\%$), it still trails OpenAIReview by $30$ points.
Reviewer3 also trails on prose errors, particularly reasoning ($10.0\%$), with its strongest category being experimental ($35.4\%$).
Per-subtype recall is in Table~\ref{tab:recall-by-type-full} in Appendix~\ref{app:perturb-full-cohort}.

%% file: analysis.tex
\section{Review Analysis}
\label{sec:analysis}

\begin{figure}[t]
    \centering
    \begin{subfigure}{0.49\columnwidth}
        \centering
        \includegraphics[width=\linewidth]{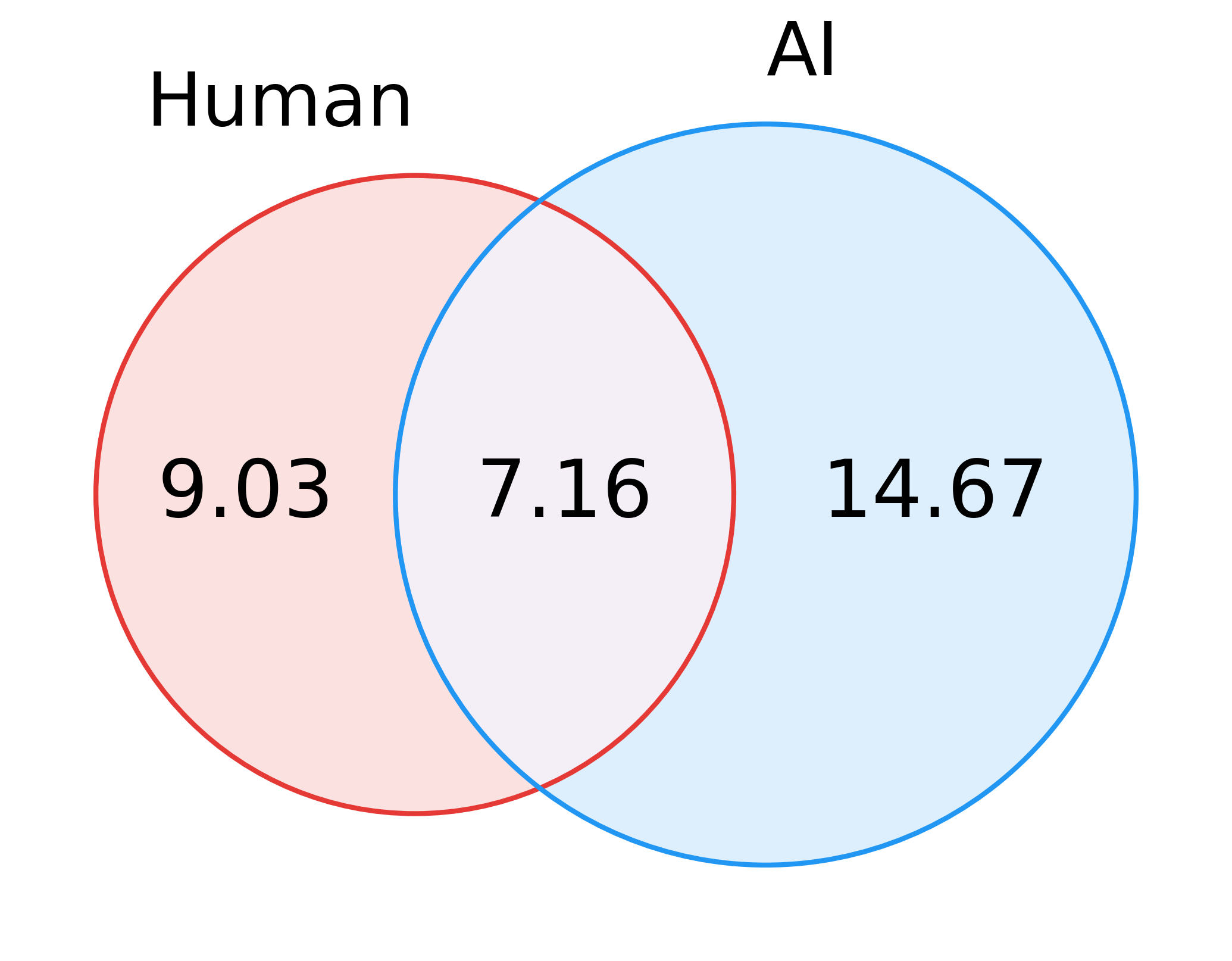}
        \caption{Across sources: human reviewers vs.\ union of the three AI systems using their best backend models (Jaccard $0.230$).}
        \label{fig:overlap_human_ai}
    \end{subfigure}
    \hfill
    \begin{subfigure}{0.49\columnwidth}
        \centering
        \includegraphics[width=\linewidth]{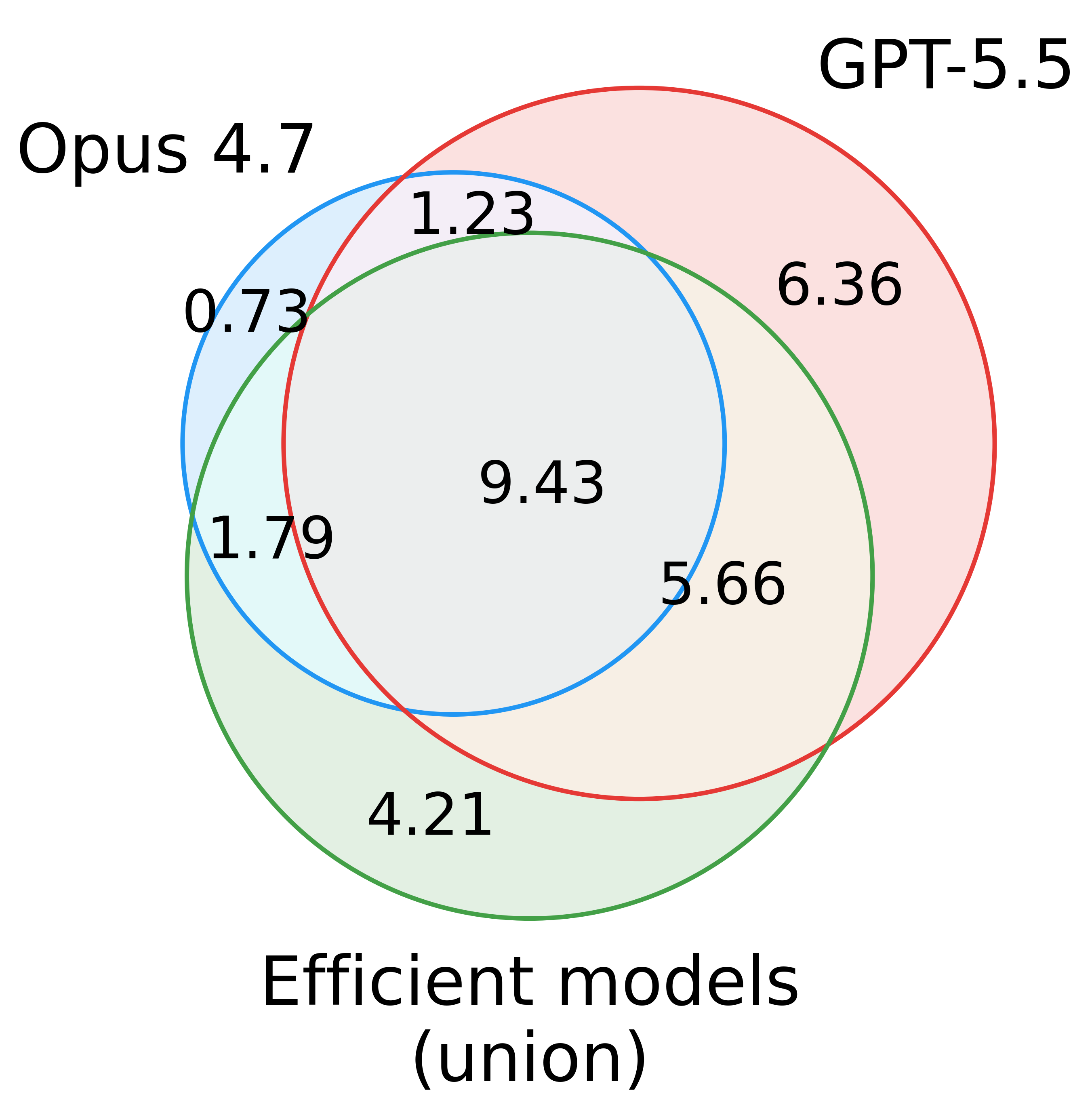}
        \caption{Across models: Claude Opus 4.7, GPT-5.5, and the union of the efficient models under OpenAIReview (3-way Jaccard $0.316$).}
        \label{fig:overlap_models}
    \end{subfigure}
    \caption{Reviewers are complementary at the paragraph level. Human reviewers and the AI union flag largely disjoint paragraphs (left), while different backend models under OpenAIReview are decently complementary (right).
    Values are average unique paragraphs per paper in each region.
    The left panel is computed on the frontier subset of papers used in Section~\ref{sec:conference}. The right panel uses the perturbation benchmark frontier subset.
    }
    \label{fig:overlap_combined}
\end{figure}

\begin{table*}[t]
    \small
    \centering
    \resizebox{\textwidth}{!}{%
    \begin{tabular}{@{}p{2.7cm} p{10.5cm} c c@{}}
    \toprule
    \textbf{Group} & \textbf{Example Comment} & \textbf{Humans \%} & \textbf{AI \%} \\
    \midrule
    \itshape Overall baseline &  & \itshape 39\% & \itshape 61\% \\
    \midrule
    Paper-Level / Cross-cutting & \emph{Human:} ``Limited novelty as it combines two existing models, closely following the approach of [1] but replacing the CNN-based sequence embedding with [2].'' & 80\% & 20\% \\
    Surface-Level (notation)    & \emph{Human:} ``The text says `$U_t$ means a number of annotations in the training dataset', but this should be $N_t$. Otherwise, what are $N_t$ and $N_s$?'' & 39\% & 61\% \\
    Claims \& Assertions        & \emph{AI:} ``The paper claims instruction tuning is `not useful' for tasks formulated as language modeling, but FLAN outperforms LaMDA-PT on 3 of 7 tasks.'' & 38\% & 62\% \\
    Experimental \& Evaluation  & \emph{AI:} ``Results in Table 2 are reported without any measure of variability or significance tests, so improvements over baselines may be due to random chance.'' & 32\% & 68\% \\
    Formulaic Math \& Derivations & \emph{AI:} ``Proposition 1's error bound depends on an undefined index $k$... and its proof is relegated to an appendix not included in the text.'' & 35\% & 65\% \\
    \bottomrule
    \end{tabular}%
    }
    \caption{Humans concentrate in paper-level / novelty critiques, while AI over-indexes on surface-level, claim, experimental, and formal-math categories. Percentages refer to who wrote the comments in each cluster: of all comments in a row's cluster, the share contributed by human reviewers (Humans \%) vs.\ the AI union (AI \%), so each row sums to $100\%$. The \emph{Overall baseline} row gives the same split over all comments pooled across clusters ($39\%$/$61\%$); a cell above its column's baseline means that source is over-represented in that cluster.
    }
    \label{tab:clusters_human_ai}
\end{table*}

Beyond aggregate recall, we want to know where in the paper each system flags issues, how they overlap or differ, and what kinds of issues those are.
We check for paragraph-level location overlap (across AI systems, models, and between AI and human reviewers) and cluster comment content to characterize how systems complement each other.

\paragraph{Method.}
For each (method, model) pair we record the set of paragraphs each reviewer commented on per paper. We present the results using Venn diagrams (Figure \ref{fig:overlap_combined}) in which each intersection and exclusive region reports the average number of paragraphs per paper, together with the Jaccard similarity of each comparison group, computed per paper and then averaged across papers.
We run this overlap analysis on both the perturbation benchmark and the quality-proxy papers of Section~\ref{sec:conference}.
To characterize the comments, we embed each comment (title and explanation) with \texttt{all-MiniLM-L6-v2} and run $k$-means clustering ($k{=}10$).
Clusters are labeled by their top TF-IDF keywords and manually merged into five interpretable groups, for which we report each source's share.
These content groups are defined independently of the injected error types and need not align with them one-to-one.
For instance, reviewers also raise issues unrelated to any perturbation, such as table or figure problems.
However, in practice the two overlap substantially.
Per-model paragraph-overlap numbers and embedding details are in Appendix~\ref{sec:review-analysis}.

\paragraph{Humans and AI are complementary in the issues they find.}
We compare the union of the three AI systems' paragraphs to those extracted from official OpenReview reviews over $70$ papers (extraction and matching details in Appendix~\ref{sec:review-analysis}).
In general, AI review systems give many more comments than humans, at 61\% versus 39\%, respectively (Table~\ref{tab:clusters_human_ai}).
Humans and AI agree on $7.16$ paragraphs per paper on average (Jaccard $0.230$), with humans raising an additional $9.03$ paper-level concerns that no AI flags, such as scope, novelty, and motivation.
In contrast, AI raises $14.67$ exclusive comments focusing more on claim validity and technical details.
Clustering all comments (Table~\ref{tab:clusters_human_ai})
shows that humans take $80\%$ of paper-level / novelty critiques (vs.\ a $39\%$ baseline), while AI leads on surface, claim, experimental, and formal-math issues, which is consistent with human reviewers triaging to high-leverage abstract concerns and AI exhaustively surfacing technical ones.
This suggests that AI review systems 
are positioned to serve peer review as a complement to human reviewers, 
handling the exhaustive enumeration of concrete issues while human reviewers focus on the higher-level judgments where they remain the dominant source of feedback.
More examples of human and AI comments from each Venn diagram region are in Table~\ref{tab:human-ai-examples} in Appendix~\ref{app:human-ai-examples}.

\paragraph{Models surface complementary errors.}
In Figure~\ref{fig:overlap_models}, under the OpenAIReview harness the backend models cover overlapping but not identical paragraphs: the union of the efficient models already covers most of what each frontier model flags---Claude Opus 4.7 leaves only $0.73$ paragraphs per paper that no other model raises, against a $9.43$-paragraph all-models intersection.
GPT-5.5 still contributes $6.36$ exclusive paragraphs per paper and efficient models contribute $4.21$.
The choice of backend also shapes \emph{what} kind of issue is raised: Claude focuses on surface-level comments while GPT leans toward claim and experimental critiques (Table~\ref{tab:clusters_claude_gpt} in Appendix~\ref{sec:review-analysis}).
Because the models catch partly different errors, collectively their detections cover $83.3\%$ of injected errors ($95\%$ CI $[79.7, 86.9]$), $11.7$ points above the best single model (GPT-5.5 at $71.6\%$).
This gain holds across a cluster bootstrap over papers ($95\%$ CI $[8.1, 15.1]$, as in Appendix~\ref{app:ci-recall}), suggesting that better harness design can potentially increase performance.
The three \emph{systems} (\coarse, OpenAIReview, Reviewer3) are likewise somewhat complementary, though less so than models (Appendix~\ref{sec:review-analysis}): OpenAIReview accounts for most flagged paragraphs, while \coarse and Reviewer3 each add only about one exclusive paragraph per paper.

%% file: user_feedback.tex
\section{User feedback}
\label{sec:production}

\subsection{Method}

We deploy the review system as a web application.
The system runs OpenAIReview with Claude Opus~4.6 as the backend model.
Users can interact with the review output in three ways: like or dislike each individual comment, click a resolve button on a comment to mark it as addressed, or leave optional free-text feedback on the review as a whole.
We analyze 1{,}360 completed reviews for 1{,}100 distinct papers.
This is an observational deployment open to the public, so the users are anonymous people on the web, most likely researchers.
The uploaded papers span many fields.\footnote{We assign fields by clustering sentence embeddings of each paper's title and abstract and labeling the clusters by inspection.}
The largest share are in computer science and AI, with the rest spread across the social sciences, life sciences, and physics.
For privacy, we do not track user information beyond these feedback signals, and we inspect paper content only as each analysis requires: titles and abstracts for the fields above, and the reviewed text behind the downvoted comments we examine below.

\subsection{Results}

\paragraph{Users vote positively and act on the comments.}
Of 27{,}587 comments shown to users, 690 received a vote  for a 2.5\% engagement rate.
Likes outnumber dislikes 407 to 283, a 1.44:1 ratio (Table~\ref{tab:votes}).
Users also marked 1{,}348 comments as resolved (4.9\% of all comments shown).
Among the 109 papers with at least one resolved comment, an average of 50\% of the comments were resolved.
We read this as promising evidence for reviews translating into author action.
The resolution rate is 24\% for upvoted comments and 41\% for downvoted ones (Table~\ref{tab:votes}).
The higher rate on downvoted comments suggests that beyond marking an issue as fixed, users also click it to dismiss a comment they disagree with.
The negative votes concentrate on specific failure modes, which we examine below.

\begin{table}[t]
  \centering
  \small
  \resizebox{\columnwidth}{!}{%
  \begin{tabular}{lcccc}
    \toprule
    \bfseries Metric & \bfseries Count & \bfseries Share & \bfseries Resolved & \bfseries Res.\ rate \\
    \midrule
    Comments shown       & 27{,}587 & 100\% & 1{,}348 & 4.9\% \\
    Votes cast           & 690      & 2.5\%  &     &       \\
    Up             & 407      & 1.5\%  & 98  & 24\%  \\
    Down           & 283      & 1.0\%  & 116 & 41\%  \\
    \bottomrule
  \end{tabular}%
  }
  \caption{Votes skew positive (1.44:1 up to down), and resolution tracking shows comments being acted on. The higher resolution rate on downvoted comments is consistent with resolution doubling as a dismissal mechanism. Comment voting and resolution over thirteen weeks of production use.}
  \label{tab:votes}
\end{table}

\paragraph{Most downvoted comments are unhelpful.}
We categorized 283 downvoted comments by the likely reason for the downvote, using an LLM judge 
and manually validated a sample of the labels (Appendix~\ref{app:downvote-errors}).
Most go to comments that are unhelpful: a false positive flags a non-issue, a trivial nitpick is valid but minor, and an unreasonable ask demands detail that the paper reasonably omits.
The full prompt given to the labeler can be found in Figure~\ref{fig:downvote-prompt} in Appendix~\ref{app:downvote-errors}.
These three modes together account for about $70\%$ of downvotes (Table~\ref{tab:downvote-errors}).
Parsing and optical character recognition (OCR) artifacts make up another $6\%$.
The rest are correct points that the author dismissed, but our manual checks reveal that some of these can also be considered minor.
The taxonomy separates distinct failure modes that all point toward the need to make the tool more targeted to substantial issues and avoid nitpicks, which is in-line with results in Section~\ref{sec:conference}.

\begin{table}[t]
  \centering
  \small
  \resizebox{\columnwidth}{!}{%
  \begin{tabular}{llrr}
    \toprule
    \bfseries Group & \bfseries Downvote reason & \bfseries $n$ & \bfseries \% \\
    \midrule
    \multirow{3}{*}{\shortstack[l]{\textbf{Unhelpful}\\$71.0\%$}}
      & False positive               & 109 & 38.5 \\
      & Trivial nitpick              & 49  & 17.3 \\
      & Unreasonable asks for detail & 43  & 15.2 \\
    \midrule
    \multirow{3}{*}{\shortstack[l]{\textbf{Other}\\$29.1\%$}}
      & Parsing / OCR artifact       & 18  & 6.4  \\
      & Correct, author dismissed    & 63  & 22.3 \\
      & Unclear                      & 1   & 0.4  \\
    \midrule
      & Total                        & 283 & 100  \\
    \bottomrule
  \end{tabular}%
  }
  \caption{Most downvoted comments are unhelpful, with false positives, nitpicks, and unreasonable detail asks together making up $71\%$. Reason each of the 283 downvoted comments was downvoted, assigned by an LLM judge (Appendix~\ref{app:downvote-errors}).}
  \label{tab:downvote-errors}
\end{table}

%% file: discussion.tex
\section{Discussion}
\label{sec:discussion}

Together, our studies show that current AI review systems already track human quality signals without additional post training, that the strongest configuration catches $71.6\%$ of injected errors, and that users of a deployed system find the comments worth acting on.
Two observations stand out.
First, in terms of raw recall, base model strength matters more than harness design: zero-shot with GPT-5.5 already reaches $59.8\%$, and the OpenAIReview harness adds $\sim$$12$ points.
The main advantage of harnesses over zero-shot is that models can output more comments, such as in the case of OpenAIReview which averages 19 comments per paper compared to zero-shot's 3.7.
Second, under the same harness, models are complementary in their comments: under OpenAIReview, the union of all models reaches $83.3\%$ recall, $11.7$ points above the best single model ($95\%$ CI $[8.1, 15.1]$). This suggests combining models via better harness design is a promising near-term direction.

Evidence from a deployed version of the system (Section~\ref{sec:production}) suggests that the capabilities our benchmarks measure are also valuable to real users.
Users who vote judge the comments helpful more often than not, and they mark comments as resolved, suggesting the reviews translate into author action.
These signals are early, and engagement rates are modest, but they point in the same direction as the benchmarks: AI review systems are already useful in practice.
Reading the downvoted comments shows where the systems fall short: most are unhelpful, raising non-issues, minor nitpicks, or over-demanding asks (Section~\ref{sec:production}).
Where the benchmarks stress recall, deployment surfaces precision as the more pressing limitation, which points to calibration and tighter prompting as concrete near-term fixes.
More generally, given that models pick up on quality signals and detect errors well, this opens up an exciting space for harness design, where systems can be tailored to different audiences or use cases, such as author-facing, area chair-facing, or domain-specific.

While we have included the best systems currently available to us, many more will likely be created in the future.
Extending the benchmark to additional systems will require standardized outputs to the format used by the evaluated systems, e.g., comments tagged with their source paragraph or span, which is only one useful design among many.
As more diverse systems are created, there will be a need for more benchmarks to measure their performance.

\section*{Limitations}
\label{sec:limitations}

Our benchmark measures recall but does not directly assess precision: a comment flagging an unperturbed passage may correspond to a pre-existing issue or a hallucinated false positive, and these cases are indistinguishable without expert annotation, which we leave to future work.
The deployment study (Section~\ref{sec:production}) gives a partial view of precision through the downvoted comments as judged by users, but it lacks expert ground truth errors on the papers themselves.
Effective reviewing also extends beyond identifying individual errors:
a useful review organizes feedback, prioritizes consequential issues, calibrates severity, and engages with the paper's broader contribution.
Thus, we view recall as a necessary but not sufficient condition for reviewer quality.
Finally, the perturbation pipeline is itself LLM-driven, which introduces a concern for bias.
The generator and verifier LLMs may share distributional biases with the LLM-based reviewers under evaluation, so the injected errors can skew toward mistakes that are salient to LLMs.
Recall measured on these perturbations could then overstate performance on the important errors that human experts would flag but that lie outside the LLM-generated distribution.
A manual audit of a stratified sample (Section~\ref{sec:perturbation}) finds $82.5\%$ of kept perturbations to be valid errors, so they are genuine mistakes, but this speaks only to their validity and not to whether they represent the errors experts most care about.
A larger audit with subject-matter experts is left to future work.

\section*{Declaration of LLM Usage}
\label{app:llm-usage}

LLMs are integral to the methodology of this paper, both as components of the benchmark pipeline and as the systems under evaluation.

\paragraph{Benchmark construction.}
The perturbation pipeline (Appendix~\ref{app:pipeline-details}) uses two LLMs in fixed roles: Gemini-3 Flash Preview as the generator that proposes candidate errors from extracted spans, and Claude Sonnet 4.6 as the checklist verifier that filters those candidates down to substantive perturbations.

\paragraph{Scoring.}
The detection-scoring stage (Appendix~\ref{app:scoring}) uses Gemini-3 Flash Preview as an explanation-matching judge that rates whether a reviewer's explanation identifies the same error as the ground-truth perturbation, on a 1--5 scale with a $\geq 3$ cutoff.

\paragraph{Systems under evaluation.}
The systems compared in this paper (OpenAIReview, \coarse, and the zero-shot baseline) are themselves LLM-based pipelines, evaluated across six backbone LLMs spanning frontier and efficient models (Section~\ref{sec:systems}).

\paragraph{Writing assistance.}
LLM-based tools were also used for editing, formatting, and prose polishing during manuscript preparation. This usage did not affect the methodology, results, or claims of the paper and is reported here only for completeness.

%% file: appendix.tex
\section{OpenAIReview Implementation Details}
\label{app:openaireview-impl}

This appendix records the configuration used by OpenAIReview (Section~\ref{sec:openaireview}) in enough detail to reproduce the harness.
All values are defaults of the reference implementation, and our experiments vary only the backbone model (Section~\ref{sec:systems}).

\paragraph{Passage segmentation.}
The extracted text is split on paragraph boundaries, and adjacent paragraphs are merged greedily into passages of up to roughly $8{,}000$ characters, so that passage lengths are comparable across papers.

\paragraph{Context window.}
When the model reviews a passage, it also sees a fixed window of neighbors: by default the five preceding and two following passages.
The asymmetry gives more weight to earlier material, where definitions and notation are introduced.

\paragraph{Running summary.}
A separate model call updates the running summary after each passage.
The summary is held to a token budget of $\max(4{,}000,\, T/10)$, where $T$ is the document's length in tokens, so the budget grows with longer papers but never falls below roughly four thousand tokens.

\paragraph{Review prompt.}
For each passage, the model receives the prompt in Figure~\ref{fig:review-prompt}: the neighboring passages and running summary as context, the passage to check, the list of checks to run, what not to flag, and the required output format.
Angle-bracketed fields are filled in at runtime, and the optical character recognition (OCR) caveat is included only when the passage text was extracted via OCR.

\begin{figure*}[t]
\begin{lstlisting}[style=prompt]
You are a thoughtful reviewer checking a passage from an academic paper. Today's date is <date>. Engage deeply with the material. For each potential issue, first try to understand the authors' intent and check whether your concern is resolved by context before flagging it.

<OCR caveat: included only when the passage text was extracted via OCR>

FULL PAPER CONTEXT (relevant sections):
<neighboring passages and the running summary of the paper so far>

---

PASSAGE TO CHECK:
<the current passage>

---

Check for:
1. Mathematical / formula errors: wrong formulas, sign errors, missing factors, incorrect derivations, subscript or index errors
2. Notation inconsistencies: symbols used in a way that contradicts their earlier definition
3. Inconsistency between text and formal definitions: prose says one thing but the equation says another
4. Parameter / numerical inconsistencies: stated values contradict what can be derived from definitions or tables elsewhere
5. Insufficient justification: a key derivation step is skipped where the result is non-trivial
6. Questionable claims: statements that overstate what has actually been shown
7. Ambiguity that could mislead: flag only if a careful reader could reasonably reach an incorrect conclusion
8. Underspecified methods: an algorithm, procedure, or modification is described too vaguely for a reader to reproduce -- key choices, boundary conditions, or parameter settings are left implicit

For each issue, write like a careful reader thinking aloud. Describe what initially confused or concerned you, what you checked to resolve it, and what specifically remains problematic. Acknowledge what the authors got right before noting the issue. Reference standard results or conventions in the field when relevant.

Be lenient with:
- Introductory and overview sections, which intentionally simplify or gloss over details
- Forward references -- symbols or claims that may be defined or justified later in the paper
- Informal prose that paraphrases a formal result without repeating every qualifier

Do NOT flag:
- Formatting, typesetting, or capitalization issues
- References to equations or sections not shown in the context (they exist elsewhere)
- Trivial observations that any reader in the field would immediately resolve
- Incomplete text at passage boundaries
- Notation not yet in the summary -- it may be introduced later

Return ONLY a JSON array (can be []). Each item:
- "title": concise title of the issue
- "quote": the exact verbatim text (preserving LaTeX)
- "explanation": deep reasoning -- what you initially thought, whether context resolves it, and what specifically remains problematic
- "type": "technical" or "logical"
\end{lstlisting}
\caption{The full prompt used to review a single passage, reproduced verbatim.
Angle-bracketed fields are substituted at runtime.}
\label{fig:review-prompt}
\end{figure*}

\paragraph{Other prompts.}
The running summary is maintained by a separate prompt (Figure~\ref{fig:summary-prompt}).
Two further steps each use a single prompt, both shown in Figure~\ref{fig:aux-prompts}: the consolidation call that merges and deduplicates the collected comments, and the opening overall-feedback paragraph, which the model writes from the first $8{,}000$ characters of the paper.

\begin{figure*}[t]
\begin{lstlisting}[style=prompt]
You are maintaining a concise running summary of an academic paper's key technical content. This summary will be used as context when reviewing later sections of the paper.

CURRENT SUMMARY:
<the running summary so far>

---

NEW PASSAGE (section <i> of <n>):
<the current passage>

---

Update the summary to incorporate any NEW information from this passage. Keep the summary structured and concise. Include:

1. **Notation & Definitions**: Any new symbols, variables, or terms defined
2. **Key Equations**: Important equations or formulas introduced (write them out, preserving LaTeX)
3. **Theorems & Propositions**: Statements of theorems, lemmas, corollaries (brief statement, not proof)
4. **Assumptions**: Any stated assumptions or conditions
5. **Key Claims**: Important results or conclusions established

Rules:
- PRESERVE all existing summary content unless it is superseded by new information
- ADD new items from the passage
- Do NOT include commentary, proof details, or experimental results
- Do NOT include information not in the passage or existing summary
- Keep entries brief -- one line per item where possible
- If the passage contains no new definitions, equations, or key claims, return the summary unchanged

Return the updated summary directly (no JSON, no code fences).
\end{lstlisting}
\caption{The prompt that updates the running summary after each passage, reproduced verbatim.
Angle-bracketed fields are substituted at runtime.}
\label{fig:summary-prompt}
\end{figure*}

\begin{figure*}[t]
\textbf{(a) Consolidation.} Merges and deduplicates the collected comments.
\begin{lstlisting}[style=prompt]
You are reviewing the complete list of issues found in an academic paper. Your job is to consolidate this list: remove duplicates and merge closely related issues.

Remove issues that:
- Flag the same underlying problem as another issue (keep the better-explained one)
- Flag standard conventions, notational shorthands, or well-known results

ISSUES FOUND:
<all collected comments, as JSON>

Return a JSON array containing the consolidated issues (same format as input). Return [] if none survive filtering.
\end{lstlisting}
\textbf{(b) Overall feedback.} Writes the opening paragraph of the review.
\begin{lstlisting}[style=prompt]
You are an expert academic reviewer. Based on the beginning of the paper below, write one paragraph of high-level feedback on the paper's quality, clarity, and most significant issues.

PAPER (first 8000 characters):
<the start of the paper>
\end{lstlisting}
\caption{The two remaining prompts in the pipeline, reproduced verbatim.
Angle-bracketed fields are substituted at runtime.}
\label{fig:aux-prompts}
\end{figure*}

\paragraph{Consolidation.}
Once every passage is reviewed, the collected comments are serialized into a single list and passed to one model call that returns a deduplicated list, merging comments that flag the same underlying issue and dropping ones that restate standard conventions.
Each surviving comment is re-anchored to its source passage by matching its quoted text.

\paragraph{Comment fields.}
Each comment carries a quoted passage, a one-line title, an explanation, a source passage index, and an optional severity (minor, moderate, or major).
Figure~\ref{fig:openaireview-example} shows an example of the resulting output.

\begin{figure*}[t]
\centering
\small
\setlength{\fboxsep}{8pt}
\fbox{\begin{minipage}{0.95\textwidth}
\textbf{Paper:} \emph{Finetuned Language Models Are Zero-Shot Learners}\\[4pt]
\textbf{Overall feedback.}
The paper is well written, clearly motivated, and presents a simple but important idea---finetuning large language models on many instruction-formatted tasks to improve zero-shot generalization---that is likely to be influential and practically useful.
The framing is strong: the contrast between standard finetuning, prompting, and instruction tuning is intuitive, and the leave-one-task-cluster-out evaluation is a reasonable first attempt to test generalization to unseen task types.
The empirical claims are compelling, especially the reported gains over the untuned 137B model and comparisons to GPT-3, and the ablation directions address key factors such as scale, number of tasks, and the role of natural-language instructions.
That said, the most significant concerns are about the strength and interpretation of the ``zero-shot'' claim: FLAN is not zero-shot in the broad sense, but rather multitask-supervised and evaluated on held-out task clusters, so the paper should be careful about possible leakage through related datasets, shared label spaces, similar input-output formats, or pretraining contamination.
The comparison to GPT-3 is also difficult to interpret because model architectures, pretraining data, prompt choices, decoding methods, and evaluation protocols may differ, making it less than a controlled comparison.
Reproducibility is another major issue, since the central experiments depend on a 137B proprietary model and a large curated instruction mixture, so full details of templates, mixture weights, dataset filtering, and prompt selection are essential.
Overall, the paper appears high quality and clearly presented, with a simple method and strong empirical results, but its main claims would be strengthened by more careful discussion of evaluation leakage, fair baselines, significance, and the precise sense in which the resulting model should be considered a zero-shot learner.

\smallskip\hrule\smallskip
\textbf{Comment 1.}\\
\emph{Quoted passage:} ``FLAN substantially improves the performance of its unmodified counterpart and surpasses zero-shot 175B GPT-3 on 20 of 25 datasets that we evaluate.''\\
\emph{Explanation:} The paper defines both mean-template and best-dev-template evaluation, but the strongest headline comparisons are not presented consistently with that qualifier: the abstract and introduction omit the best-dev-template caveat, while nearby figures use mean-template FLAN results.
[\ldots] Since template choice can materially affect zero-shot performance, readers could mistake the 20-of-25 claims for fully template-free zero-shot results.

\smallskip\hrule\smallskip
\textbf{Comment 2.}\\
\emph{Quoted passage:} ``Translation (8 datasets): ParaCrawl EN/DE, EN/ES, EN/FR; WMT-16 EN/CS, EN/DE, EN/FI, EN/RO, EN/RU, EN/TR.''\\
\emph{Explanation:} The figure lists three ParaCrawl language pairs plus six WMT-16 language pairs, which would total nine translation datasets, while the header says eight.
This may be a formatting problem, or it may reflect a convention about what counts as one dataset, but as written it creates ambiguity about the exact instruction-tuning mixture and the stated total dataset count.

\smallskip\hrule\smallskip
\textbf{Comment 3.}\\
\emph{Quoted passage:} ``performance improvements from instruction tuning emerge only with sufficient model scale.''\\
\emph{Explanation:} The section reports a useful model-size ablation, but it does not estimate a scaling law in the usual sense: no functional relationship is fit, no uncertainty or extrapolation analysis is provided, and the evidence comes from one model family and one instruction-tuning recipe.
The conclusion that improvements emerge ``only'' with sufficient scale therefore risks overgeneralizing; a more precise phrasing would say that, in these experiments, gains appeared only for the larger tested models.
\end{minipage}}
\caption{An example OpenAIReview output, with one paragraph of overall feedback followed by comments, each pairing a quoted passage with an explanation. ``\ldots'' marks elisions in the longer comments.}
\label{fig:openaireview-example}
\end{figure*}

\paragraph{Document parsing.}
Review quality depends on clean text, since math symbols, tables, and reading order are easy to corrupt during extraction.
PDFs are converted through a fallback chain of parsers, tried in order until one succeeds: a cloud OCR service (Mistral OCR), a local OCR model (DeepSeek OCR), a layout-preserving local parser (Marker), and finally an offline extractor (pymupdf4llm).
LaTeX, Word, and arXiv HTML inputs bypass OCR and are converted directly.

\section{Correlation with Human Quality Signals}
\label{app:conference-full-cohort}

\subsection{Sampling design}
\label{app:conference-sampling}

We randomly subsample within the selection criteria so that papers in each proxy minimally overlap.
This applies to both groups of the Conference label and to the rejected group of the Community label. The other groups are selected deterministically by their ranking criterion.
By design, none of the four labels is a uniform sample of the conference papers.
Each group is a deliberately selected contrast meant to surface a quality signal.

\paragraph{Quality-proxy construction.}
All signals come from the SNOR records (Semantic Scholar citation counts, OpenReview decisions, publication venue, and review scores). We compute \emph{citations-per-year} as the citation count divided by the number of years since the paper's conference year. The four proxies' groups are then:
\textbf{Community}: high: the top papers by citations-per-year (deterministic, ties broken by forum id); low (``never published''): papers that were not accepted, have $\ge 3$ reviews, are $\ge 2$ years past their conference year, and have no formal publication venue (empty or arXiv-only), sampled at random.
\textbf{Conference}: high: papers whose normalized decision matches an award keyword (Outstanding/Best/Oral/Spotlight); low: not-accepted papers with $\ge 3$ reviews; both sampled at random.
\textbf{Reviewer}: the top and bottom papers by mean review score among papers with $\ge 3$ reviews.
\textbf{Composite}: high: awarded \emph{and} top-cited \emph{and} top-scored; low: rejected \emph{and} never-published \emph{and} bottom-scored.
Random samples use a fixed seed for reproducibility, and every group requires $\ge 3$ official reviews (our ``substantive-reviews'' filter).

\paragraph{Frontier subset cost.}
Running the two frontier models (GPT-5.5 and Claude Opus 4.7) under the zero-shot and OpenAIReview methods on the 80-paper subset cost approximately \$204 in API charges (\$177 for OpenAIReview and \$27 for zero-shot). Extending the frontier sweep to the full 240-paper set would multiply this proportionally, which is why we restrict frontier runs to the subset.

\paragraph{Review accounting.}
The benchmark evaluates 197 unique papers $\times$ 4 efficient models $\times$ 3 methods $= 2{,}364$ reviews on the full 240-paper set, plus 74 unique papers $\times$ 2 frontier models $\times$ 2 methods (zero-shot and OpenAIReview, no \coarse) $= 296$ reviews on the 80-paper frontier subset, for $2{,}660$ reviews in total.

\subsection{Interpreting the pairwise accuracy}
\label{app:auc-interpretation}

The main text (Section~\ref{sec:conference}) reports the \emph{pairwise accuracy} (equivalently, the Mann--Whitney AUC~\citep{mannwhitney1947,hanley1982auc}).
Concretely, for each quality proxy we form every (low-group paper, high-group paper) pair and count it as a hit if the low-group paper receives more comments than the high-group one (ties contribute $0.5$). Pairwise accuracy is the hit fraction.
Aggregating across the four proxies gives $400$ pairs in the frontier subset and $3{,}600$ pairs in the full-set appendix tables.

Three intuitions help anchor the numbers:

\begin{itemize}[nosep]
    \item \textbf{Probabilistic reading.} The pairwise accuracy is exactly the probability that, if you sample one low-group paper and one high-group paper uniformly at random within the same proxy, the low-group paper will have received more comments, under random tiebreaking when the two papers have identical counts (the $0.5$ contribution per tie is the expected outcome of that coin flip).
    A pairwise accuracy of $0.83$ (GPT-5.5) means that this happens $83\%$ of the time.
    \item \textbf{Chance baseline.} Under the null hypothesis that comment volume is independent of paper quality, the expected pairwise accuracy is $0.5$. Values $> 0.5$ mean the system tracks the proxy direction. Values $< 0.5$ mean it inverts it.
    \item \textbf{$\Delta$ vs.\ pairwise accuracy.} $\Delta$ measures the average gap between the two groups' means and is sensitive to outlier papers with very high comment counts. Pairwise accuracy measures the per-pair ordering and is robust to those outliers. The two metrics usually agree on direction but can disagree in magnitude.
\end{itemize}

\subsection{Confidence intervals for pairwise accuracy}
\label{app:ci-accuracy}

All pairwise-accuracy intervals reported in Section~\ref{sec:conference} are 95\% intervals from a nonparametric \emph{cluster bootstrap} over papers.
The same paper appears in many (low, high) pair comparisons across the four quality proxies (and across the per-severity tiers), so those pairs are not independent draws of comments.
A naive interval that treated each pair as an independent Bernoulli trial would overstate precision.
For each (method, model) cell, each bootstrap draw works as follows: within each proxy we form a new high-quality group by drawing papers \emph{with replacement} from the original high-quality papers (so some papers appear two or three times and others not at all, but the group size stays the same), and similarly form a new low-quality group. We then recompute the pooled AUC on these resampled groups.
Per-severity tier AUCs (Appendix~\ref{app:conference-tier-breakdown}) reuse the same per-bootstrap resample. Repeating this $B = 5000$ times gives the bootstrap distribution of the AUC; we report its $2.5$th and $97.5$th percentiles.

\subsection{Results on the full 240-paper set}

The main-text conference tables (Section~\ref{sec:conference}) restrict all six models to the 80-paper frontier subset (74 unique). Tables~\ref{tab:conference-model-aggregate-full} and~\ref{tab:conference-severity-aggregate-full} report the same breakdowns under OpenAIReview on the full 240-paper set (197 unique).

\begin{table}[t]
  \centering
  \resizebox{\columnwidth}{!}{%
  \begin{tabular}{lccccc}
    \toprule
    \bfseries Model & \bfseries $\bar{c}_{\rm high}$ & \bfseries $\bar{c}_{\rm low}$ & \bfseries $\Delta$ & \bfseries \% $\Delta$ & \bfseries Acc. \\
    \midrule
    Grok-4.1-Fast                 & 2.66  & 6.93  & $\mathbf{4.27}$ & $\mathbf{160.3\%}$ & \ci{$\mathbf{0.81}$}{0.75, 0.86} \\
    DeepSeek-V4-Flash             & 12.18 & 14.44 & $2.26$          & $18.5\%$           & \ci{$0.61$}{0.54, 0.68} \\
    Qwen3.6-35B-A3B               & 11.70 & 13.45 & $1.75$          & $15.0\%$           & \ci{$0.62$}{0.55, 0.69} \\
    Gemini-3.1-Flash-Lite         & 13.10 & 14.16 & $1.06$          & $8.1\%$            & \ci{$0.62$}{0.54, 0.69} \\
    \midrule
    Reviewer3                     & 6.20  & 9.04  & $2.84$          & $45.8\%$           & \ci{$0.79$}{0.73, 0.84} \\
    \bottomrule
  \end{tabular}%
  }
  \caption{Under OpenAIReview, every efficient model and Reviewer3 emit more comments on low-quality papers than high-quality ones on the full 240-paper set (197 unique), with Grok-4.1-Fast attaining the highest pairwise accuracy ($0.81$). Brackets are 95\% bootstrap CIs over papers (Appendix~\ref{app:ci-accuracy}).}
  \label{tab:conference-model-aggregate-full}
\end{table}

\paragraph{The signal concentrates in the major and moderate tiers.}
Decomposing per-model pairwise accuracy by severity tier (Table~\ref{tab:conference-severity-aggregate-full}) shows that the quality signal concentrates in the major and moderate tiers. The minor tier, omitted from the table for consistency with the per-system tier breakdown in Appendix~\ref{app:conference-tier-breakdown}, sits at or below chance for several of the efficient models (DeepSeek $0.47$, Qwen $0.48$, Gemini $0.34$).

\begin{table}[t]
  \centering
  \resizebox{\columnwidth}{!}{%
  \begin{tabular}{lcccc}
    \toprule
    \bfseries Model & \bfseries $\bar{c}_{\rm l}$ & \bfseries Maj & \bfseries Mod & \bfseries Overall \\
    \midrule
    Grok-4.1-Fast                 & 6.93  & \ci{$0.64$}{0.59, 0.68} & \ci{$\mathbf{0.79}$}{0.73, 0.84} & \ci{$\mathbf{0.81}$}{0.75, 0.86} \\
    DeepSeek-V4-Flash             & 14.44 & \ci{$0.68$}{0.62, 0.74} & \ci{$0.66$}{0.59, 0.72} & \ci{$0.61$}{0.54, 0.68} \\
    Qwen3.6-35B-A3B               & 13.45 & \ci{$0.59$}{0.53, 0.65} & \ci{$0.72$}{0.65, 0.78} & \ci{$0.62$}{0.55, 0.69} \\
    Gemini-3.1-Flash-Lite         & 14.16 & \ci{$0.62$}{0.56, 0.67} & \ci{$0.77$}{0.71, 0.82} & \ci{$0.62$}{0.54, 0.69} \\
    \midrule
    Reviewer3                     & 9.04  & \ci{$\mathbf{0.78}$}{0.72, 0.83} & \ci{$0.67$}{0.60, 0.73} & \ci{$0.79$}{0.73, 0.84} \\
    \bottomrule
  \end{tabular}%
  }
  \caption{Under \textbf{OpenAIReview}, per-tier pairwise accuracy concentrates in the major and moderate tiers on the full 240-paper set (197 unique). The minor tier is omitted (see prose); $\bar{c}_{\rm l}$ is the mean number of comments on the low-quality group; brackets are 95\% bootstrap CIs over papers (Appendix~\ref{app:ci-accuracy}).}
  \label{tab:conference-severity-aggregate-full}
\end{table}

\subsection{Severity-tier pairwise accuracy}
\label{app:conference-tier-breakdown}

The main paper reports overall pairwise accuracy only. We retain the per-tier breakdown here as a robustness check, separately for each system (zero-shot, OpenAIReview, \coarse) across the six backbone models on the 74-paper frontier subset.
We omit the minor tier due to it being a much noisier signal for quality compared to major or moderate.
Tables~\ref{tab:conference-zeroshot-tier},~\ref{tab:conference-progressive-tier}, and~\ref{tab:conference-coarse-tier} show that the rank order of models is largely preserved across tiers, but the Major and Moderate tiers do not always agree on the per-model ordering.
Bracketed quantities are 95\% bootstrap CIs over papers (Appendix~\ref{app:ci-accuracy}).

\begin{table}[t]
  \centering
  \resizebox{\columnwidth}{!}{%
  \begin{tabular}{lcccc}
    \toprule
    \bfseries Model & \bfseries $\bar{c}_{\rm l}$ & \bfseries Maj & \bfseries Mod & \bfseries Overall \\
    \midrule
    GPT-5.5                       & 7.25 & \ci{0.74}{0.65, 0.82} & \ci{0.77}{0.67, 0.86} & \ci{0.78}{0.68, 0.87} \\
    Claude Opus 4.7               & 9.25 & \ci{0.69}{0.61, 0.76} & \ci{0.84}{0.75, 0.91} & \ci{0.79}{0.69, 0.88} \\
    DeepSeek-V4-Flash             & 4.75 & \ci{$\mathbf{0.78}$}{0.69, 0.86} & \ci{0.71}{0.61, 0.80} & \ci{$\mathbf{0.80}$}{0.71, 0.89} \\
    Qwen3.6-35B-A3B               & 3.45 & \ci{0.60}{0.53, 0.68} & \ci{0.56}{0.44, 0.68} & \ci{0.59}{0.46, 0.71} \\
    Gemini-3.1-Flash-Lite         & 2.92 & \ci{0.58}{0.53, 0.62} & \ci{$\mathbf{0.87}$}{0.80, 0.94} & \ci{0.76}{0.64, 0.86} \\
    Grok-4.1-Fast                 & 3.25 & \ci{0.63}{0.55, 0.70} & \ci{0.73}{0.63, 0.82} & \ci{0.74}{0.63, 0.84} \\
    \bottomrule
  \end{tabular}%
  }
  \caption{Under \textbf{zero-shot} on the frontier subset, DeepSeek-V4-Flash leads overall ($0.80$) and Gemini-3.1-Flash-Lite has the highest moderate-tier accuracy ($0.87$). $\bar{c}_{\rm l}$ is the mean number of comments on the low-quality group.}
  \label{tab:conference-zeroshot-tier}
\end{table}

\begin{table}[t]
  \centering
  \resizebox{\columnwidth}{!}{%
  \begin{tabular}{lcccc}
    \toprule
    \bfseries Model & \bfseries $\bar{c}_{\rm l}$ & \bfseries Maj & \bfseries Mod & \bfseries Overall \\
    \midrule
    GPT-5.5                       & 22.15 & \ci{$\mathbf{0.78}$}{0.69, 0.86} & \ci{$\mathbf{0.91}$}{0.83, 0.96} & \ci{$\mathbf{0.83}$}{0.73, 0.91} \\
    Claude Opus 4.7               & 14.62 & \ci{0.59}{0.54, 0.64} & \ci{0.86}{0.78, 0.93} & \ci{0.74}{0.63, 0.83} \\
    Grok-4.1-Fast                 & 6.80  & \ci{0.63}{0.56, 0.70} & \ci{0.75}{0.65, 0.85} & \ci{0.80}{0.69, 0.89} \\
    DeepSeek-V4-Flash             & 13.95 & \ci{0.65}{0.55, 0.76} & \ci{0.61}{0.48, 0.73} & \ci{0.60}{0.47, 0.71} \\
    Qwen3.6-35B-A3B               & 14.43 & \ci{0.59}{0.49, 0.69} & \ci{0.72}{0.60, 0.82} & \ci{0.62}{0.49, 0.74} \\
    Gemini-3.1-Flash-Lite         & 14.12 & \ci{0.66}{0.57, 0.75} & \ci{0.79}{0.68, 0.88} & \ci{0.61}{0.48, 0.74} \\
    \bottomrule
  \end{tabular}%
  }
  \caption{Under \textbf{OpenAIReview} on the frontier subset, GPT-5.5 leads every tier (Major $0.78$, Moderate $0.91$, Overall $0.83$). $\bar{c}_{\rm l}$ is the mean number of comments on the low-quality group.}
  \label{tab:conference-progressive-tier}
\end{table}

\begin{table}[t]
  \centering
  \resizebox{\columnwidth}{!}{%
  \begin{tabular}{lcccc}
    \toprule
    \bfseries Model & \bfseries $\bar{c}_{\rm l}$ & \bfseries Maj & \bfseries Mod & \bfseries Overall \\
    \midrule
    DeepSeek-V4-Flash             & 19.32 & \ci{$\mathbf{0.72}$}{0.60, 0.84} & \ci{0.63}{0.50, 0.75} & \ci{0.60}{0.47, 0.72} \\
    Qwen3.6-35B-A3B               & 15.80 & \ci{0.57}{0.47, 0.67} & \ci{0.69}{0.57, 0.80} & \ci{0.66}{0.54, 0.78} \\
    Gemini-3.1-Flash-Lite         & 5.67  & \ci{0.62}{0.53, 0.70} & \ci{0.63}{0.50, 0.74} & \ci{0.58}{0.46, 0.70} \\
    Grok-4.1-Fast                 & 7.30  & \ci{0.69}{0.60, 0.78} & \ci{$\mathbf{0.71}$}{0.60, 0.81} & \ci{$\mathbf{0.66}$}{0.54, 0.78} \\
    \bottomrule
  \end{tabular}%
  }
  \caption{Under \textbf{\coarse} on the frontier subset, Grok-4.1-Fast attains the best overall ($0.66$) and moderate-tier ($0.71$) accuracy. GPT-5.5 and Claude Opus 4.7 were not run on \coarse. $\bar{c}_{\rm l}$ is the mean number of comments on the low-quality group.}
  \label{tab:conference-coarse-tier}
\end{table}

\subsection{OpenAIReview: raw vs.\ consolidated output}
\label{app:conference-raw-vs-cons}

OpenAIReview emits two comment lists: the \emph{raw} pre-consolidation output (one comment per passage, before deduplication) and the \emph{consolidated} output (after the post-hoc consolidation step that merges duplicates and re-tiers comments).
The main text uses the raw output throughout. Here we report the comparison.
On DeepSeek and Qwen, the consolidated lists average $9.3$ and $10.1$ comments per paper, respectively, against $13.3$ and $12.6$ raw, roughly a $20$--$30\%$ reduction.
Per-proxy $\Delta$ on the proxies values shift modestly: on DeepSeek, raw $\{3.57, 0.55, 1.80, 3.10\}$ vs.\ consolidated $\{2.34, 0.93, 1.70, 3.13\}$ (community, conference, reviewer, and composite). On Qwen, raw $\{3.55, 1.97, 0.80, 0.67\}$ vs.\ consolidated $\{1.63, 0.07, 0.57, 0.33\}$, with consolidation tending to reduce the gap, especially on Qwen.

\section{Perturbation Benchmark}
\label{app:benchmark-creation}

\subsection{Full perturbation taxonomy with examples}

Table~\ref{tab:error-taxonomy-full} summarizes the taxonomy of injected perturbations, spanning surface-level edits and higher-level claim, logical, and experimental errors.

\begin{table*}[t]
  \centering
  \small
  \resizebox{\textwidth}{!}{%
  \begin{tabular}{
    >{\raggedright\arraybackslash}p{1.8cm}
    >{\raggedright\arraybackslash}p{3.6cm}
    >{\raggedright\arraybackslash}p{10.5cm}
  }
    \toprule
    \bfseries Category & \bfseries Subtype & \bfseries Example \\
    \midrule
    \multirow{4}{1.8cm}{Surface}
      & Operator / sign      & $+ \to -$, \: $\leq \to \geq$, \: $\cup \to \cap$ \\[2pt]
      & Index / subscript    & $x_i \to x_{i+1}$, \: $A^n \to A^{n-1}$ \\[2pt]
      & Numeric              & $0.5 \to 0.25$, \: $n=100 \to n=10$ \\[2pt]
      & Computation          & $2+3=3$; \: $\frac{1-\rho}{\rho} \to \frac{\rho}{1-\rho}$ \\[2pt]
    \midrule
    \multirow{2}{1.8cm}{Claim}
      & False theoretical claim & Dropping a domain restriction (``If $f$ is continuous on a \emph{compact} set, then $f$ is bounded'' $\to$ ``If $f$ is continuous, then $f$ is bounded'') \\[2pt]
      & False empirical claim   & Inverting a standard empirical fact (``Lower mean squared error (MSE) indicates better predictive performance'' $\to$ ``Higher mean squared error (MSE) indicates better predictive performance.'') \\[2pt]
    \midrule
    \multirow{4}{1.8cm}{Logic}
      & Circular reasoning   & The proof concludes that $f$ is injective because it has an inverse, but the existence of the inverse is justified by assuming $f$ is injective \\[2pt]
      & Invalid implication  & If $ab = 0$, then $a = 0$ \\[2pt]
      & Induction error      & Skipping the inductive step or using an incorrect base case \\[2pt]
      & Missing case         & The argument considers $x > 0$ and concludes the result holds for all $x$, ignoring the case $x \le 0$ \\[2pt]
    \midrule
    \multirow{3}{1.8cm}{Experimental}
      & Reversed causality   & ``Increasing the sample size reduces the variance of the estimator'' $\to$ ``Lower variance in the estimator causes an increase in sample size'' \\[2pt]
      & Misinterpretation of results & ''A $p$-value of $0.20$ provides strong evidence against the null hypothesis'' \\[2pt]
      & P-hacking            & ``We repeated the experiment under different random seeds and report the run with the strongest performance.'' \\[2pt]
    \bottomrule
  \end{tabular}%
  }
  \caption{Taxonomy of injected perturbations with examples. Surface errors are local math-token edits. Claim, reasoning, and experimental errors are paragraph-level edits that introduce a more abstract error. Examples are loosely based on perturbations in our benchmark.}
  \label{tab:error-taxonomy-full}
\end{table*}

\subsection{Error-type to span mapping}

Table~\ref{tab:error-span} defines the span types used in the perturbation pipeline and the corresponding eligible error categories.

\begin{table*}[t]
  \centering
  \small
  \resizebox{\textwidth}{!}{%
  \begin{tabular}{p{4.0cm}p{3.0cm}p{8.0cm}}
    \toprule
    \bfseries Span Type & \bfseries Paper Type & \bfseries Admissible Error Categories \\
    \midrule
    Equation                    & Theoretical, Empirical & Surface: operator/sign, index/subscript, numeric, computation \\[3pt]
    Definition / Theorem        & Theoretical            & Claim: false theoretical claim \\[3pt]
    Proof                       & Theoretical            & Reasoning: circular, induction, invalid implication, missing case \\[3pt]
    Claim / Argument paragraph  & Empirical              & Claim: false theoretical or empirical claim \\[3pt]
    Experimental paragraph      & Empirical              & Experimental: reversed causality, misinterpretation of results, p-hacking \\
    \bottomrule
  \end{tabular}%
  }
  \caption{Span types and the error categories admissible for each. Math-token edits (Surface) target equations. Prose-level edits (Claim, Logic, Experimental) target spans appropriate to the paper type.}
  \label{tab:error-span}
\end{table*}

\subsection{Pipeline-stage implementation details}
\label{app:pipeline-details}

The main-text Section~\ref{sec:perturbation} summarizes the five-stage pipeline (\textit{extract}, \textit{generate}, \textit{validate}, \textit{verify}, \textit{inject}). We give the per-stage details here.

\paragraph{Extract.}
A deterministic \LaTeX{}-source scanner finds places where errors can be injected. For theoretical papers, it extracts equations, definitions/theorems, and proofs. For empirical papers, it extracts equations, paragraphs that make claims or arguments, and paragraphs describing experimental setup or results. These candidate spans are mapped to allowed perturbations according to Table~\ref{tab:error-span}.

\paragraph{Generate.}
A generator LLM is first prompted with the abstract to identify the paper's field and write a short list of plausible field-specific errors, which is appended to the main generation prompt as guidance. Each candidate's input record includes the span's \LaTeX{} text, its type (display/inline/named equation, definition/theorem, proof, or paragraph), a $\pm 200$-character window of surrounding prose for context, and the set of error subtypes compatible with that span type. The generator is shown all candidates and selects a subset, writing one perturbation per chosen candidate. For each, it emits (i)~the specific error label (one of the admissible subtypes), (ii)~the replacement \LaTeX{} text, (iii)~a short explanation of how a careful reader could verify the error from the paper alone, and (iv)~optionally, a verbatim quote from elsewhere in the paper that the perturbation directly contradicts. Candidates are batched and the generator is asked to aim for 20 valid perturbations per paper. We use Gemini-3 Flash Preview as the perturbation generator.

\paragraph{Validate.}
A structural validation step rejects perturbations where (i)~the perturbed text is identical to the original, (ii)~the span overlaps with an already-accepted perturbation, or (iii)~the replacement would create garbled \LaTeX{} at the span boundaries. Duplicate perturbations on the same span are also dropped here.

\paragraph{Verify.}
Surviving perturbations next pass through a \emph{checklist verifier} that filters out edits too local to constitute a substantive error or those that are technically not errors.
To do this, we find other passages in the paper related to the perturbed span and check if it contradicts them.
Related passages are computed by tokenizing the span (\LaTeX{} commands, scripted identifiers like $W_{ij}$ or $x_{t+1}$, variable assignments, capitalized noun phrases) and searching the rest of the document for hits, returning up to five $\pm 100$-character snippets around the matches.
A deterministic precheck first rejects structurally typo-shaped edits without any model call, such as a dummy-variable rename or a symbol swap whose replacement letter is bound nowhere in the paper.
The remaining perturbations from one run (one paper and one error family) are then judged in a single batched call, with the verifier instructed to judge each perturbation independently.
Each perturbation's record contains the original span, the perturbed span, the error label, a $\pm 200$-character window of surrounding prose, and the span's related passages.
For each perturbation, the verifier answers a 4-item yes/no checklist and returns a short verbatim quote from the perturbed text pinpointing the issue.
We use Claude~Sonnet~4.6 as the verifier.

The answer pattern maps deterministically to one of three verdicts:
\begin{itemize}[nosep]
  \item if item~1 = N \emph{or} item~4 = Y $\Rightarrow$ \emph{typo-shaped} (reject).
  \item else if item~2 = N $\Rightarrow$ \emph{not-an-error} (reject).
  \item else if item~3 = N $\Rightarrow$ \emph{not-an-error} (reject).
  \item else $\Rightarrow$ \emph{substantive} (keep).
\end{itemize}
Item~1 and item~4 are local well-formedness / cosmetic-edit checks that pre-empt typo-shaped artifacts. Item~2 verifies that the available evidence pins down the original content. Item~3 asks whether the perturbation breaks something concrete that the evidence relies on. The checklist is shared across error families except item~3, whose contradiction criterion is specific to each family (Surface, Reasoning, Claim, Empirical). The items are:

\begin{enumerate}[nosep]
  \item \emph{Well-formed:} Is the perturbed span well-formed when read in isolation (no mixed-direction inequalities, broken sandwich expressions, operator salad, type/unit mismatches, garbled grammar, unbound symbols, or definitions with undefined symbols, missing required quantifiers, or mismatched arity)?
  \item \emph{Evidence available:} Is there a concrete basis to judge the perturbation against: either the surrounding context or related passages establish something specific about the same object (a stated value, definition, applied theorem, named result, or downstream use of the same quantity), or the perturbed text alone introduces a self-evidencing methodological flaw (e.g., post-hoc selection, a removed multiple-testing correction, treating a p-value as the probability of the null)? Mentioning the topic abstractly does not count.
  \item \emph{Contradiction confirmed:} Does the perturbation contradict that evidence, judged per error family? Surface: the perturbed value, symbol, or operator now disagrees with how the same quantity appears elsewhere. Reasoning: the perturbed step breaks the chain of inference (a missing case whose conclusion is not forced by the remaining cases, an induction whose base case is now wrong or whose step no longer reduces $n+1$ to $n$, a step that now invokes the very claim being proved, or a reversed/dropped implication). Claim: the altered definition or theorem no longer supports an application of it visible in the evidence, e.g., weakened/strengthened quantifiers ($\forall \leftrightarrow \exists$) or a dropped/reversed hypothesis the application uses. Empirical: the perturbed claim disagrees with the established content (a number/dataset/method contradiction, a misread of what a result means, a flipped causal direction) or introduces a methodological flaw the original explicitly avoided.
  \item \emph{Typo-shaped or cosmetic:} Is the change typo-shaped or cosmetically equivalent regardless of evidence (a bare symbol swap whose replacement letter is not bound anywhere in the paper, a re-indexing that leaves the expression algebraically unchanged such as the dummy-variable rename $\sum_i \to \sum_j$, synonym swaps, reordering equivalent clauses, renaming bound variables, or hedging tweaks that do not flip the conclusion)?
\end{enumerate}

\paragraph{Inject.}
The previous stages produce a list of approved (span, replacement) records but leave the source paper untouched. Injection applies all surviving edits to the \LaTeX{} source: because a replacement may change the length of the text, perturbations are sorted by offset in descending order and applied right-to-left, so each replacement only modifies text past positions that have already been finalized and the remaining (lower-offset) perturbations keep the offsets recorded during extraction without any recalculation. The result is a single corrupted paper carrying all of its accepted perturbations.

\paragraph{Manual validation.}
Beyond the automated validate/verify filters, one of the authors manually audited a stratified sample of $40$ kept perturbations to confirm they are genuine, well-formed errors. The sample takes $10$ perturbations per error type (Surface, Claim, Logic, Experimental), drawn so that every subtype appears at least once and all eight domains are represented, and spread across papers (fixed random seed). For each perturbation the annotator was shown the original span, the injected replacement, the generator's why-wrong explanation, and the verifier's contradicting evidence, and labeled it a \emph{valid error}, \emph{not an error}, or \emph{ambiguous}. We judge $33/40$ ($82.5\%$) to be valid, $2$ ($5\%$) not true errors, and $5$ ($12.5\%$) ambiguous, with per-type valid rates of Surface $8/10$, Claim $7/10$, Logic $9/10$, and Experimental $9/10$. The non-valid and ambiguous cases concentrate in two subtypes. (i)~\emph{Surface-numeric} edits where the extracted span bounds only part of a number, so injection can produce a malformed expression, e.g., replacing the ``$\times$'' token of ``$5\times10^{-6}$'' yields the duplicated exponent ``$5\times10^{-12}10^{-6}$'' rather than a clean magnitude change. (ii)~\emph{Empirical-claim} edits whose contradiction with the rest of the paper is not concretely pinned down by an identifiable passage, so the injected statement is unsupported rather than provably wrong. These cases motivate the larger expert audit we leave to future work (Section~\ref{sec:limitations}).

\subsection{Generator span selection vs.\ random selection}
\label{app:selection-bias}

Because the generator chooses which candidate spans to perturb, the injected errors could be biased toward spans the model prefers, relative to random selection from the same candidates.
Since extraction is deterministic, we can reconstruct the exact candidate pool each generation run saw and compare the chosen spans against that pool; under random selection the two distributions would match.
We run this audit over all 222 generation runs (74 papers, one run per error family).
Reconstruction reproduces every pool exactly, and all injected perturbations map back to their source spans verbatim.
Selection is measured at the earliest recorded stage, after structural validation, which retains 3{,}577 of 3{,}670 generated perturbations (97.5\%).

\paragraph{Selection freedom is limited.}
In 96 of 222 runs (43\%), the generator perturbed every candidate it was shown, leaving no room for selection bias; the analysis below uses the remaining runs.
Moreover, candidates are presented in batches of 10 that are nearly always homogeneous in span type, with a fixed perturbation target per batch, so the allocation of errors across span types is largely set by the pipeline rather than chosen by the model.

\paragraph{Where the generator's preferences show.}
For equation spans, selection rate rises with span length, from 22\% in the shortest within-run quartile to 44\% in the longest, so the generator favors substantial equations over short inline math.
Named equation environments are selected somewhat more often than inline math (40\% vs.\ 33\%; rate difference $0.08$, 95\% CI $[0.03, 0.13]$, bootstrap over papers).
Equations later in the paper are mildly favored (29\% in the earliest within-run quartile vs.\ 43\% in the latest).
Among the admissible equation subtypes, the generator prefers operator/sign edits (37\%) over index/subscript (27\%), numeric (23\%), and computation (13\%) edits.
We find no position-in-prompt bias: a within-batch permutation test shows no preference for candidates listed earlier or later in the prompt ($p \ge 0.18$ for every error family).

\subsection{Corpus details}
\label{app:corpus}

The benchmark draws on 74 recent arXiv papers spanning eight subject classes: cs.CC, cs.LG, econ.EM, hep-ex, math.* (covering math.AG, math.CO, math.NT, math.PR, and math.ST), physics.atm-clus, q-bio.GN, and stat.AP.
We aim for 10 papers per class and accept 5 from physics.atm-clus and 9 from math.* due to availability and \LaTeX{} compilation constraints.
The perturbation pipeline yields 29--60 retained perturbations per paper (median 46) for a total of 3{,}365 injected edits after validation, verification, and deduplication.
Per-cell denominators in the recall tables (Section~\ref{sec:perturbation}, Appendix~\ref{app:perturb-full-cohort}) are smaller than these raw counts because they restrict to the (model, method) cells that completed scoring. We report the exact scored counts in each table cell.

\begin{table}[t]
  \centering
  \resizebox{\columnwidth}{!}{%
  \begin{tabular}{lrrrrrr}
    \toprule
    \bfseries Domain & \bfseries Papers & \bfseries Surface & \bfseries Claim & \bfseries Reasoning & \bfseries Experimental & \bfseries Total \\
    \midrule
    cs.CC             & 10 & 141 & 136 & 128 & --   & 405 \\
    cs.LG             & 10 & 160 & 180 & --  & 194  & 534 \\
    econ.EM           & 10 & 139 & 139 & --  & 173  & 451 \\
    hep-ex            & 10 & 129 & 145 & --  & 160  & 434 \\
    math.*            &  9 & 129 & 117 &  93 & --   & 339 \\
    physics.atm-clus  &  5 &  65 &  65 & --  &  71  & 201 \\
    q-bio.GN          & 10 & 123 & 176 & --  & 188  & 487 \\
    stat.AP           & 10 & 156 & 165 & --  & 193  & 514 \\
    \midrule
    \bfseries Total         & \bfseries 74 & \bfseries 1{,}042 & \bfseries 1{,}123 & \bfseries 221 & \bfseries 979 & \bfseries 3{,}365 \\
    \bottomrule
  \end{tabular}%
  }
  \caption{The corpus is balanced across domains, with reasoning perturbations concentrated in cs.CC and math.* and experimental perturbations in the empirical sciences. Counts are retained perturbations after validation, verification, and deduplication. ``Claim'' aggregates false theoretical claims (perturbed definitions and theorem statements) and false empirical claims (perturbed prose statements). ``Reasoning'' errors are injected only into proofs, so they appear only in the proof-heavy classes.}
  \label{tab:corpus-by-domain}
\end{table}

\subsection{Scoring details}
\label{app:scoring}

The substring-match stage normalizes whitespace and capitalization and requires that the perturbed text cover at least 75\% of the comment's quoted span (or vice versa).
The LLM judge is Gemini-3 Flash Preview. For each comment passing the substring stage it rates the explanation against the ground truth on a 1--5 scale, and any rating $\geq 3$ counts as a detection.
A perturbation is counted as detected if any emitted comment passes both stages.

\paragraph{Threshold sensitivity.}
Detection recall is nearly flat around the chosen 0.75: moving $\tau$ anywhere in $[0.6, 0.9]$ changes recall by at most one detection out of 58 (25 detections for $\tau \le 0.75$, 24 above), so results are not sensitive to the threshold (Table~\ref{tab:threshold-sweep}).
Lowering $\tau$ to $0.5$ admits roughly 3$\times$ more candidate pairs into the LLM-judge stage but recovers only two additional detections, suggesting the looser threshold mainly admits noisy near-matches. Raising $\tau$ above 0.9 begins to reject otherwise-valid detections where the reviewer paraphrases the perturbed string.
The sweep is run on the math domain with Claude-Opus-4.7 as the reviewer (58 perturbations across the OpenAIReview and zero-shot methods) and uses Gemini-3 Flash Preview as the explanation judge, with judge results cached per (perturbation, comment) pair so each pair is judged once and re-aggregated at every threshold.

\begin{table}[t]
  \centering
  \small
  \resizebox{\columnwidth}{!}{%
  \begin{tabular}{rrrr}
    \toprule
    $\tau$ & pairs passing substring & detected & recall \\
    \midrule
    0.50 & 377 & 27 & 0.466 \\
    0.60 & 227 & 25 & 0.431 \\
    0.70 & 164 & 25 & 0.431 \\
    0.75 & 138 & 25 & 0.431 \;\;(default) \\
    0.80 & 115 & 24 & 0.414 \\
    0.85 & 102 & 24 & 0.414 \\
    0.90 &  88 & 24 & 0.414 \\
    0.95 &  79 & 23 & 0.397 \\
    1.00 &  78 & 23 & 0.397 \\
    \bottomrule
  \end{tabular}%
  }
  \caption{Detection recall changes by at most one detection (of 58) anywhere in $\tau \in [0.6, 0.9]$, supporting the 0.75 default. ``Pairs passing'' counts (perturbation, comment) pairs whose normalized coverage is at least $\tau$. ``Detected'' counts perturbations for which at least one such pair also passes the LLM judge. Sweep slice: math papers reviewed by Claude-Opus-4.7, $n=58$ injected perturbations.}
  \label{tab:threshold-sweep}
\end{table}

\subsection{Confidence intervals for recall}
\label{app:ci-recall}

All recall intervals reported in Section~\ref{sec:perturbation} are 95\% intervals from a nonparametric \emph{cluster bootstrap} over papers.
Two perturbations injected into the same paper share the paper's writing style, domain, and exposition, so a reviewer that catches one is more likely to catch the others; they are not statistically independent draws.
A naive interval that treated each perturbation as an independent Bernoulli trial would overstate precision.
For each (method, model) cell, each bootstrap draw works as follows: we form a new set of papers by drawing \emph{with replacement} from the scored papers (so some papers appear two or three times and others not at all, but the total number of papers stays the same), and recompute the pooled recall $\sum_p \mathrm{detected}_p / \sum_p \mathrm{injected}_p$ on that resample. Repeating $B = 5000$ times gives the bootstrap distribution; we report its $2.5$th and $97.5$th percentiles.
Intervals are correspondingly wide for cells backed by few papers (e.g., the Reasoning category, present only in cs.CC and math).

\subsection{Perturbation benchmark results}
\label{app:perturb-full-cohort}

The main-text recall tables in Section~\ref{sec:perturbation} all restrict to the 24-paper frontier subset, so that the frontier-model rows (Claude-Opus-4.7 and GPT-5.5) and the four efficient-model rows share the same papers. Tables~\ref{tab:recall-overall-full}, \ref{tab:recall-by-type-full}, and~\ref{tab:recall-by-domain-full} below report parallel breakdowns on the full 74-paper benchmark, aggregated across the four efficient models that ran on every paper, as a robustness check on the frontier-subset main-text view.

\paragraph{OpenAIReview outperforms zero-shot and \coarse on nearly all domains.}
Table~\ref{tab:recall-by-domain} reports recall by paper domain on the 24-paper frontier subset, using each system's best-performing backend (GPT-5.5 for zero-shot and OpenAIReview, DeepSeek-V4-Flash for \coarse).
The ordering OpenAIReview $>$ zero-shot $>$ \coarse holds in every domain except Math, where zero-shot with the same GPT-5.5 backend leads OpenAIReview (79.3\% vs.\ 62.1\%).
Empirical-science domains (Econometrics 91.8\%, ML 84.7\%, Atomic Phys.\ 72.0\%, HEP 71.8\%) reach the highest OpenAIReview recall, while the theoretical Complexity domain lags at 45.7\%.

\begin{table*}[t]
  \centering
  \small
  \resizebox{\textwidth}{!}{%
  \begin{tabular}{llccccccccc}
    \toprule
    \bfseries Method & \bfseries Backend & \bfseries Overall & \bfseries Econometrics & \bfseries ML & \bfseries Atomic Phys. & \bfseries HEP & \bfseries Genomics & \bfseries Statistics & \bfseries Math & \bfseries Complexity \\
    \midrule
    zero-shot     & GPT-5.5     & 59.8\%          & 75.3\%          & 67.2\%          & 65.3\%          & 67.3\%          & 58.9\%          & 45.0\%          & \textbf{79.3\%} & 33.7\%          \\
    \coarse       & DeepSeek-V4 & 20.7\%          & 32.9\%          & 32.1\%          & 15.3\%          & 19.1\%          & 20.9\%          & 15.6\%          & 20.7\%          & 8.7\%           \\
    OpenAIReview  & GPT-5.5     & \textbf{71.6\%} & \textbf{91.8\%} & \textbf{84.7\%} & \textbf{72.0\%} & \textbf{71.8\%} & \textbf{71.3\%} & \textbf{66.1\%} & 62.1\%          & \textbf{45.7\%} \\
    Reviewer3     & closed      & 26.5\%          & 27.7\%          & 23.4\%          & 32.2\%          & 12.8\%          & 50.6\%          & 29.4\%          & 0.0\%           & 18.5\%          \\
    \bottomrule
  \end{tabular}%
  }
  \caption{OpenAIReview tops every arXiv domain except Math, where zero-shot with the same backend leads. Per-domain columns ordered by OpenAIReview recall (descending). All cells use the 24-paper frontier subset (where GPT-5.5 and Claude-Opus-4.7 were also run), with each system's best-performing backend. Reviewer3 has no model selector, and a few of its runs on Econometrics, HEP, and Genomics did not complete scoring.}
  \label{tab:recall-by-domain}
\end{table*}

\paragraph{Recall on the full set preserves the main-text ordering.}
Aggregated across the four efficient models on all 74 papers, the system ordering OpenAIReview $>$ zero-shot $>$ \coarse holds without exception: by model (Table~\ref{tab:recall-overall-full}), by error type (Table~\ref{tab:recall-by-type-full}), and by domain (Table~\ref{tab:recall-by-domain-full}).
OpenAIReview improves on zero-shot for every efficient backend, with the largest margin on DeepSeek-V4-Flash ($29.4\% \to 55.4\%$), and \coarse trails zero-shot in every cell, mirroring the frontier-subset result in the main text.
The error-type and domain breakdowns echo the main-text findings: OpenAIReview's gains are largest on the prose-level categories (claim $48.6\%$, reasoning $54.4\%$, experimental $48.5\%$ overall) and on the empirical-science domains (cs.LG $54.0\%$, econ.EM $53.2\%$), while the theoretical cs.CC domain remains the hardest ($32.7\%$).
Absolute recall is lower than in the main-text tables because the full set omits the two frontier models, on which every system scores highest.

\begin{table}[t]
  \centering
  \small
  \resizebox{\columnwidth}{!}{%
  \begin{tabular}{lccc}
    \toprule
    \bfseries Model & \bfseries \coarse & \bfseries zero-shot & \bfseries OpenAIReview \\
    \midrule
    Grok-4.1-Fast                 & \cnt{14.8\%}{343/2{,}320}  & \cnt{33.5\%}{782/2{,}337}    & \cnt{\textbf{49.3\%}}{1{,}153/2{,}337} \\
    DeepSeek-V4-Flash             & \cnt{21.2\%}{490/2{,}307}   & \cnt{29.4\%}{686/2{,}337}    & \cnt{\textbf{55.4\%}}{1{,}259/2{,}274} \\
    Qwen3.6-35B-A3B               & \cnt{15.9\%}{371/2{,}337}   & \cnt{31.7\%}{740/2{,}337}    & \cnt{\textbf{46.2\%}}{1{,}079/2{,}337} \\
    Gemini-3.1-Flash-Lite         & \cnt{12.1\%}{282/2{,}337}    & \cnt{14.3\%}{334/2{,}337}     & \cnt{\textbf{30.9\%}}{723/2{,}337} \\
    \bottomrule
  \end{tabular}%
  }
  \caption{OpenAIReview beats zero-shot and \coarse on every efficient backend on the full 74-paper benchmark. 
  }
  \label{tab:recall-overall-full}
\end{table}

\begin{table}[t]
  \centering
  \small
  \resizebox{\columnwidth}{!}{%
  \begin{tabular}{lccc}
    \toprule
    \bfseries Error type & \bfseries \coarse & \bfseries zero-shot & \bfseries OpenAIReview \\
    \midrule
    Operator / sign     & \cnt{36.4\%}{168/461} & \cnt{45.4\%}{339/746} & \cnt{\textbf{56.1\%}}{530/944} \\
    Index / subscript   & \cnt{31.6\%}{97/307}     & \cnt{38.8\%}{209/539}     & \cnt{\textbf{49.1\%}}{335/682} \\
    Numeric parameter   & \cnt{40.3\%}{85/211}     & \cnt{47.6\%}{161/338}     & \cnt{\textbf{57.1\%}}{268/469} \\
    Computation         & \cnt{57.9\%}{70/121}      & \cnt{61.1\%}{116/190}     & \cnt{\textbf{68.3\%}}{181/265} \\
    Claim               & \cnt{18.1\%}{528/2{,}910}  & \cnt{29.6\%}{949/3{,}211}  & \cnt{\textbf{48.6\%}}{1{,}672/3{,}443} \\
    Reasoning           & \cnt{29.1\%}{39/134}      & \cnt{38.1\%}{43/113}        & \cnt{\textbf{54.4\%}}{80/147} \\
    Experimental        & \cnt{21.4\%}{499/2{,}337}   & \cnt{31.2\%}{725/2{,}321}   & \cnt{\textbf{48.5\%}}{1{,}148/2{,}365} \\
    \midrule
    \bfseries Overall         & \cnt{16.0\%}{1{,}486/9{,}301}  & \cnt{27.2\%}{2{,}542/9{,}348} & \cnt{\textbf{45.4\%}}{4{,}214/9{,}285} \\
    \bottomrule
  \end{tabular}%
  }
  \caption{OpenAIReview wins on every error type on the full 74-paper benchmark; the prose-level categories (Claim, Reasoning, Experimental) show the largest absolute gains over zero-shot. Each cell pools the four efficient backends (Grok-4.1-Fast, DeepSeek-V4-Flash, Qwen3.6-35B-A3B, and Gemini-3.1-Flash-Lite) under that system; the frontier models are excluded. Denominators differ across systems because only runs that completed scoring are counted.}
  \label{tab:recall-by-type-full}
\end{table}

\begin{table*}[t]
  \centering
  \small
  \begin{tabular}{lccc}
    \toprule
    \bfseries Domain & \bfseries \coarse & \bfseries zero-shot & \bfseries OpenAIReview \\
    \midrule
    Mathematics (math.*)                            & 16.1\% {\scriptsize (58/360)}      & 29.7\% {\scriptsize (107/360)}       & \textbf{45.8\%} {\scriptsize (165/360)} \\
    Computational Complexity (cs.CC)                & 10.1\% {\scriptsize (75/743)}      & 17.4\% {\scriptsize (131/752)}      & \textbf{32.7\%} {\scriptsize (246/752)} \\
    Experimental High-Energy Physics (hep-ex)       & 17.4\% {\scriptsize (225/1{,}293)}   & 28.3\% {\scriptsize (368/1{,}300)}  & \textbf{46.0\%} {\scriptsize (590/1{,}282)} \\
    Econometrics (econ.EM)                          & 21.0\% {\scriptsize (229/1{,}092)}   & 34.0\% {\scriptsize (371/1{,}092)}  & \textbf{53.2\%} {\scriptsize (566/1{,}063)} \\
    Applied Statistics (stat.AP)                    & 9.2\% {\scriptsize (136/1{,}472)}  & 19.4\% {\scriptsize (285/1{,}472)}  & \textbf{38.3\%} {\scriptsize (564/1{,}472)} \\
    Genomics (q-bio.GN)                             & 15.5\% {\scriptsize (282/1{,}823)}  & 24.1\% {\scriptsize (444/1{,}840)}  & \textbf{44.2\%} {\scriptsize (814/1{,}840)} \\
    Atomic and Cluster Physics (physics.atm-clus)   & 16.4\% {\scriptsize (126/770)}       & 27.8\% {\scriptsize (218/784)}      & \textbf{42.3\%} {\scriptsize (325/768)} \\
    Machine Learning (cs.LG)                        & 20.3\% {\scriptsize (355/1{,}748)}  & 35.4\% {\scriptsize (618/1{,}748)}   & \textbf{54.0\%} {\scriptsize (944/1{,}748)} \\
    \bottomrule
  \end{tabular}
  \caption{OpenAIReview wins on every domain on the full 74-paper benchmark, with the strongest leads on cs.LG (54.0\%) and econ.EM (53.2\%); cs.CC remains the hardest (32.7\%).}
  \label{tab:recall-by-domain-full}
\end{table*}

\section{Review Analysis}
\label{sec:review-analysis}

\subsection{Embedding and clustering details}
\label{app:embedding-details}

We characterize what each method talks about by clustering comment text in a shared embedding space.
We embed every comment with the \texttt{sentence-transformers/all-MiniLM-L6-v2} encoder (384-d, default pooling) and run k-means with $k=10$ and \texttt{random\_state=42} on the raw embeddings.
For interpretability we fit a TF-IDF vectorizer (\texttt{max\_features=10000}, English stop-words) on the same texts and label each cluster with its top-15 average-TF-IDF terms. We additionally surface the five comments closest to each centroid as exemplars.
Two of the authors then merge the 10 raw clusters into 5 interpretable groups by inspecting keywords and exemplars.
Each comment is concatenated as \texttt{title + " " + explanation} before embedding.

We run two clustering passes.
The first (Appendix~\ref{app:overlap-cp}) clusters \coarse and OpenAIReview comments from the efficient models over all 197 papers, totalling 12{,}704 comments (7{,}526 \coarse + 5{,}178 OpenAIReview).
The second (Appendix~\ref{app:overlap-all}) adds zero-shot, totalling 14{,}456 comments (7{,}526 + 5{,}178 + 1{,}752).
Per-paper comment volumes are uneven across models and methods: \coarse averages 18.1 comments/paper for DeepSeek and 14.7 for Qwen but only 5.4 for Gemini, while OpenAIReview is more uniform (6.0--10.1 comments/paper across models).
This volume gap matters when reading the overlap statistics below, because Jaccard scales with set size.

\subsection{Across-systems paragraph overlap}
\label{app:overlap-systems}

\begin{figure}[t]
    \centering
    \includegraphics[width=0.85\columnwidth]{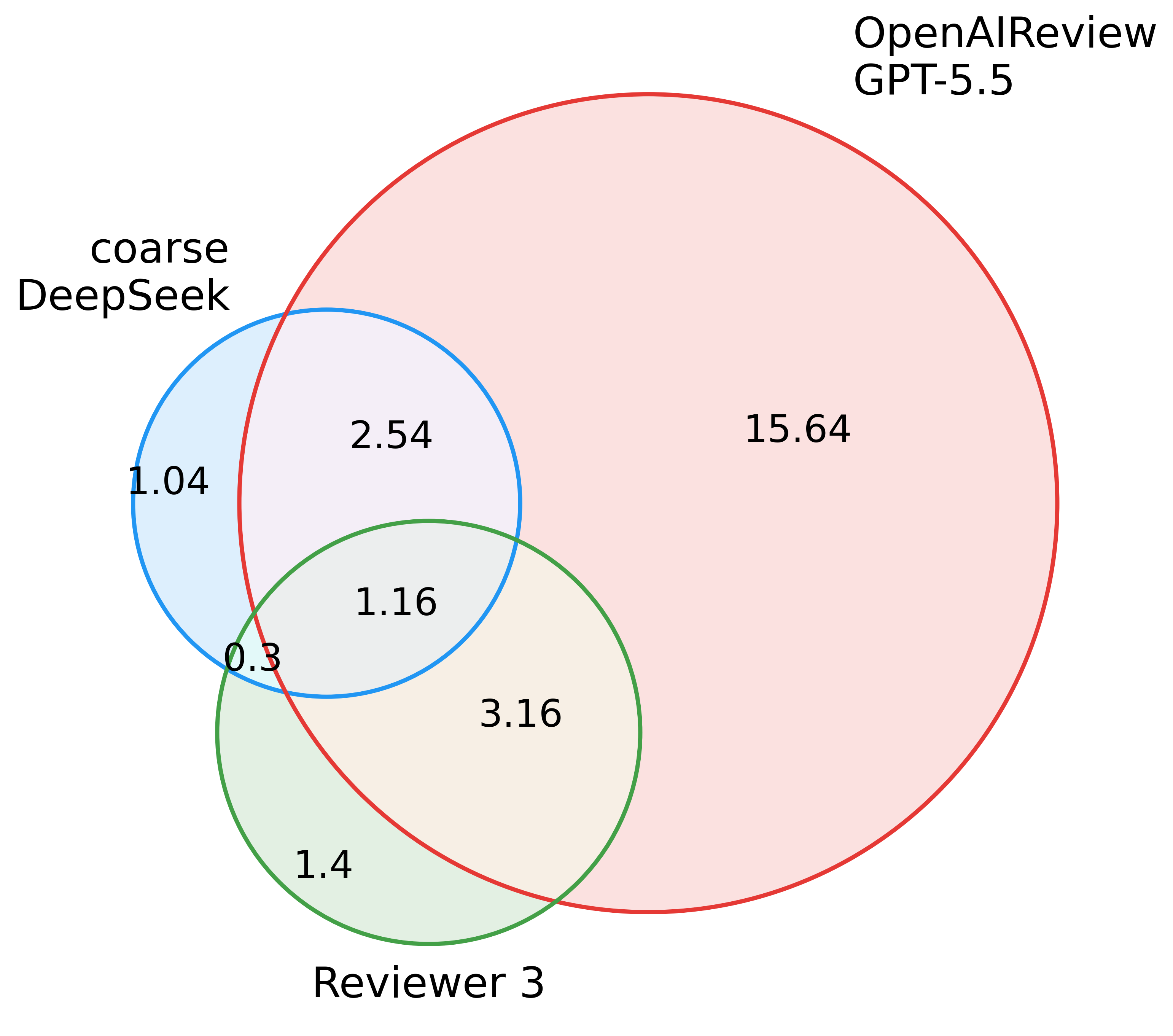}
    \caption{Across systems on the perturbation benchmark frontier subset, using each system's best backend (\coarse~/ DeepSeek-V4-Flash, OpenAIReview~/ GPT-5.5, Reviewer3): the three systems target largely disjoint paragraphs (3-way Jaccard $0.046$). Values are average unique paragraphs per paper in each region.}
    \label{fig:overlap_three_systems}
\end{figure}

\paragraph{The three AI systems target different paragraphs (Figure~\ref{fig:overlap_three_systems}).}
Across all papers in the frontier subset and all error types,
the three-way intersection averages only $1.16$ paragraphs per paper out of $\approx 25$ paragraphs touched by at least one system (3-way Jaccard $0.046$).
OpenAIReview and Reviewer3 overlap the most among pairs ($4.3$ shared paragraphs per paper), while \coarse contributes only $1.0$ paragraph per paper that no other system raises.
In terms of volume, OpenAIReview emits the most comments, taking up $69\%$ of all comments vs.\ $13\%$ for \coarse and $19\%$ for Reviewer3, and once normalized against this baseline no system shows a clear specialization in their comment content (Table~\ref{tab:clusters_cp}).
Looking at the type of comments, the systems mostly agree on surface notation and formal-math issues but diverge on the higher-level categories: \coarse leans toward claim-related critiques and Reviewer3 toward experimental and statistical issues, while OpenAIReview spreads more evenly across categories.

\begin{figure}[t]
    \centering
    \includegraphics[width=0.85\columnwidth]{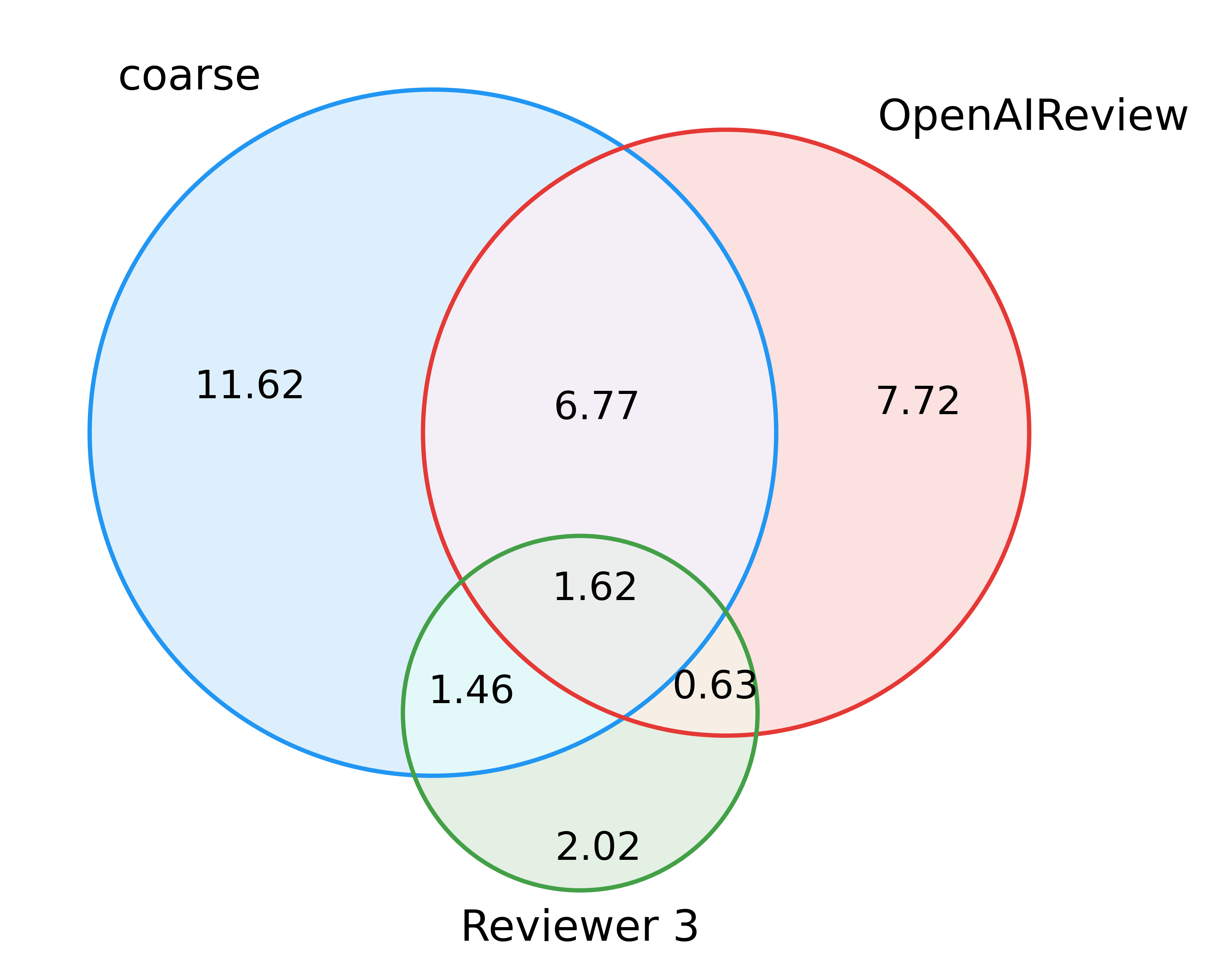}
    \caption{Across systems on the quality-proxy papers, aggregating each system over its backbone models (\coarse and OpenAIReview each unioned over the three efficient models run on the full set, while Reviewer3 has no model selector). Unioning over models enlarges every region relative to the best-model view (Figure~\ref{fig:overlap_three_systems}) but the systems still occupy largely distinct paragraphs (3-way Jaccard $0.053$). Values are average unique paragraphs per paper in each region.}
    \label{fig:overlap_union_models}
\end{figure}

\paragraph{Aggregating over all models per system (Figure~\ref{fig:overlap_union_models}).}
Pinning one best model per system understates each system's footprint.
Taking the union of paragraphs over each system's backbone models (the three efficient models each for \coarse and OpenAIReview on the quality-proxy papers, with Reviewer3 unchanged) enlarges every region but leaves the qualitative picture intact: \coarse and OpenAIReview share the largest pairwise region ($6.8$ paragraphs/paper) yet the three-way intersection is still only $1.6$ paragraphs/paper (3-way Jaccard $0.053$), confirming that the systems address complementary slices of each paper rather than converging as models are added.

\subsection{\coarse and OpenAIReview}
\label{app:overlap-cp}

The two methods comment on largely disjoint sets of paragraphs: average per-paper Jaccard ranges from 0.106 (Gemini) to 0.164 (Qwen), and intersection sizes are small in absolute terms (Table~\ref{tab:overlap_cp}).
DeepSeek and Qwen flag many \coarse-only paragraphs (10.5 and 8.9 per paper) because they are the highest-volume \coarse models. Even so, the shared region is at most 2.9 paragraphs per paper.
Across all three models, fewer than 20\% of flagged paragraphs are shared, indicating that \coarse and OpenAIReview consistently address different issues.

\begin{table}[t]
  \centering
  \resizebox{\columnwidth}{!}{%
  \begin{tabular}{lcccc}
    \toprule
    Model & \coarse only & OpenAIReview only & Both & Jaccard \\
    \midrule
    DeepSeek-V4-Flash       & 10.54 & 5.34 & 2.41 & 0.129 \\
    Qwen3.6-35B-A3B         &  8.87 & 5.66 & 2.88 & 0.164 \\
    Gemini-3.1-Flash-Lite   &  4.19 & 5.15 & 1.05 & 0.106 \\
    \bottomrule
  \end{tabular}%
  }
  \caption{\coarse and OpenAIReview flag largely disjoint paragraphs across all three efficient models. Per-paper averages over 197 papers. ``\coarse only'' and ``OpenAIReview only'' count paragraphs flagged by exactly one method, ``Both'' counts the intersection, and Jaccard is averaged across papers.}
  \label{tab:overlap_cp}
\end{table}

\subsection{\coarse, OpenAIReview, and zero-shot}
\label{app:overlap-all}

Adding zero-shot leaves the picture essentially unchanged: three-way agreement is rare and zero-shot contributes almost nothing that the other two methods do not already cover (Table~\ref{tab:overlap_all}).
The three-way intersection peaks at 0.58 paragraphs per paper for DeepSeek.
Zero-shot's volume is small to begin with (1.0--3.3 paragraphs/paper) and most of it lands outside the \coarse$\cap$OpenAIReview region, so the three-way Jaccard collapses below 0.03 for every model.
The largest pairwise intersection that excludes zero-shot (\coarse$\cap$OpenAIReview only) is 1.8--2.4 paragraphs per paper for the high-volume DeepSeek and Qwen, mirroring the two-way result above.
Together with Appendix~\ref{app:overlap-cp}, this confirms that the three methods address complementary slices of each paper rather than rediscovering the same issues.

\begin{table*}[t]
  \centering
  \small
  \setlength{\tabcolsep}{4pt}
  \begin{tabular}{lcccccccc}
    \toprule
    Model & C only & O only & Z only & C$\cap$O only & C$\cap$Z only & O$\cap$Z only & C$\cap$O$\cap$Z & Jaccard \\
    \midrule
    DeepSeek-V4-Flash      & 9.85 & 4.61 & 1.30 & 1.83 & 0.70 & 0.73 & 0.58 & 0.028 \\
    Qwen3.6-35B-A3B        & 8.38 & 4.93 & 1.10 & 2.36 & 0.49 & 0.73 & 0.52 & 0.028 \\
    Gemini-3.1-Flash-Lite  & 4.01 & 4.47 & 1.06 & 0.84 & 0.18 & 0.69 & 0.21 & 0.021 \\
    \bottomrule
  \end{tabular}
  \caption{Three-way paragraph-level overlap is sparse: zero-shot adds little beyond what \coarse and OpenAIReview already cover, and the three-way intersection never exceeds 0.6 paragraphs/paper. Per-paper averages over 197 papers. C = \coarse, O = OpenAIReview, Z = zero-shot. ``C only'' counts paragraphs flagged only by \coarse, ``C$\cap$O only'' counts paragraphs flagged by both \coarse and OpenAIReview but not zero-shot, etc.}
  \label{tab:overlap_all}
\end{table*}

\subsection{Cluster breakdowns referenced from Section~\ref{sec:analysis}}
\label{app:main-cluster-tables}

Tables~\ref{tab:clusters_cp} and~\ref{tab:clusters_claude_gpt} give the cluster-share breakdowns referenced from the results discussion in Section~\ref{sec:analysis}; the humans-vs-AI breakdown (Table~\ref{tab:clusters_human_ai}) is in the main text.

\begin{table*}[t]
\centering
\resizebox{\textwidth}{!}{
\begin{tabular}{llccc}
\toprule
\textbf{Group} & \textbf{Representative Keywords and Phrases} & \textbf{\coarse \%} & \textbf{OpenAIReview \%} & \textbf{Reviewer3 \%} \\
\midrule
\itshape Overall baseline & \itshape (share of all comments) & \itshape 13\% & \itshape 69\% & \itshape 19\% \\
\midrule
Surface-Level (notation) & notation, equation, matrix, mathbf, dimension, undefined & 11\% & 70\% & 19\% \\
Claims \& Assertions & claim, overstates, contradiction, attention, architecture & 19\% & 61\% & 20\% \\
Experimental \& Evaluation & baseline, comparison, training, statistical, variance, GPU-hours & 11\% & 63\% & 25\% \\
Formulaic Math \& Derivations & bound, theorem, proof, convergence, manifold, equivariance & 15\% & 66\% & 19\% \\
Tables \& Figures & table, caption, figure, row, column, baseline & 14\% & 64\% & 22\% \\
\bottomrule
\end{tabular}
}
\caption{On the perturbation frontier subset ($50$ corrupted papers, one per source paper and error family), OpenAIReview emits the majority of comments across every cluster and within a few points of its overall share ($69\%$). No system shows a sharp topic specialization. Each row sums to $\sim 100\%$ across the three systems. The overall comment-volume baselines are $13/69/19$.}
\label{tab:clusters_cp}
\end{table*}

\begin{table*}[t]
\centering
\resizebox{\textwidth}{!}{
\begin{tabular}{lllcc}
\toprule
\textbf{Group} & \textbf{Representative Keywords and Phrases} & \textbf{Claude \%} & \textbf{GPT \%} \\
\midrule
\itshape Overall baseline & \itshape (share of all comments) & \itshape 33\% & \itshape 67\% \\
\midrule
Surface-Level & index, sign, form, inconsistency & 64\% & 36\% \\
Claims \& Assertions & claim, too broad, mischaracterized & 21\% & 79\% \\
Experimental \& Evaluation & evidence, baseline, reported, performance, validation, test & 22\% & 78\% \\
Formulaic Math \& Derivations & bound, lemma, proof, theorem, convergence & 46\% & 54\% \\
Table \& Figures & table, figure & 49\% & 51\% \\
\bottomrule
\end{tabular}
}
\caption{Relative to the $33/67$ Claude/GPT overall baseline, Claude over-indexes sharply on Surface-Level ($64\%$) and GPT over-indexes on Claims and Experimental ($79\%$, $78\%$). Formal math and tables/figures track the baseline. Each row sums to $100\%$ across the two models.}
\label{tab:clusters_claude_gpt}
\end{table*}

\subsection{Example human and AI comments by overlap region}
\label{app:human-ai-examples}

To illustrate the categorical split reported in Table~\ref{tab:clusters_human_ai}, Table~\ref{tab:human-ai-examples} shows representative comments in three regions: paragraphs flagged by both humans and at least one AI system, paragraphs flagged only by humans, and paragraphs flagged only by AI. The intersection and AI-only examples are drawn from one paper (ICLR 2021, ``Autoregressive Entity Retrieval''). The human-only examples are drawn from two other ICLR 2021 papers in our set, chosen so that the comments fall in the \emph{paper-level / cross-cutting} cluster (the cluster where humans most over-index against the $39\%$ baseline, see Table~\ref{tab:clusters_human_ai}). Even in the intersection region, humans and AI tend to comment on different aspects of the same paragraph: humans raise broader concerns about scope or methodology, while AI surfaces local notation, claim, or terminology issues.

\begin{table*}[t]
\centering
\small
\renewcommand{\arraystretch}{1.15}
\resizebox{\textwidth}{!}{%
\begin{tabular}{@{}p{1.7cm}p{2cm}p{12.8cm}@{}}
\toprule
\bfseries Region & \bfseries Source & \bfseries Comment \\
\midrule
\multirow{4}{1.7cm}{\itshape Intersection (humans \& AI)}
& Human & \emph{Impact of lack of contrastive learning}: ``one effect of contrastive learning [\dots] on systems with explicit entity representations is some implicit training of \emph{all} entities, which is lacking in the described autoregressive proposal.'' \\
& AI (OpenAIReview) & \emph{Inconsistency between `orders of magnitude' and `$\sim$20 times'}: The passage claims the memory footprint is `orders of magnitude smaller' (implying at least $100\times$ reduction) but a few sentences later states `$\sim$20 times smaller'. \\
\cmidrule(l){2-3}
& Human & \emph{Inference efficiency vs.\ MIPS retrieval}: ``Compared to the models with a large entity memory whose retrieval is performed with Maximum inner product search, how is the efficiency of your decoding strategy?'' \\
& AI (OpenAIReview) & \emph{Grammatical number agreement error}: The phrase `an established approximate decoding strategies' contains a grammatical error. `Strategies' should be singular to agree with `an'. \\
\midrule
\multirow{2}{1.7cm}{\itshape Human-only \\ (paper-level)}
& Human & \emph{Unclear distinction between novel contributions and existing work}: ``I struggled to disentangle the novel contributions of the authors from the work they were reviewing. [\dots] The authors need to do a better job at highlighting their own contributions but also making clear what is not.'' \\
& Human & \emph{Lack of novelty in using RNN/LSTM for PDEs}: ``Using the RNN/LSTM type of networks for time series / time-dependent PDEs doesn't seem to be novel. Convolution-LSTM has been widely used in these tasks.'' \\
\midrule
\multirow{2}{1.7cm}{\itshape AI-only}
& AI (OpenAIReview) & \emph{Implicit assumption of parameter sharing in cold-start claim}: The claim that the model handles `cold-start' scenarios without retraining is overstated. The model's ability to zero-shot generate new entities is limited by its BART-based sub-word compositionality. \\
& AI (\coarse) & \emph{Incorrect metric terminology for KILT results}: The KILT benchmark and entity retrieval literature standardly report Recall@$k$ rather than Precision. Claiming an improvement in `precision points' is likely a typo for `recall points'. \\
\bottomrule
\end{tabular}%
}
\caption{Representative human and AI comments in three overlap regions. Intersection rows pair each human comment with the AI comment grounded to the same paragraph. In both cases, the AI comment addresses a local notation or grammar issue while the human raises a broader methodological concern. Human-only comments are drawn from the \emph{paper-level / cross-cutting} cluster (novelty, contributions, related-work positioning). AI-only comments are substantive but localized (cold-start overclaim, terminology error). Intersection and AI-only rows are from ``Autoregressive Entity Retrieval'' (ICLR 2021). Human-only rows are from ``Unpacking Information Bottlenecks'' and ``Neural Time-Dependent PDE'' (both ICLR 2021). Excerpts are lightly abridged for length.}
\label{tab:human-ai-examples}
\end{table*}

\subsection{Quality-proxy comment overlap}
\label{app:outcomes-overlap}

\begin{figure}[t]
\centering
\includegraphics[width=0.85\columnwidth]{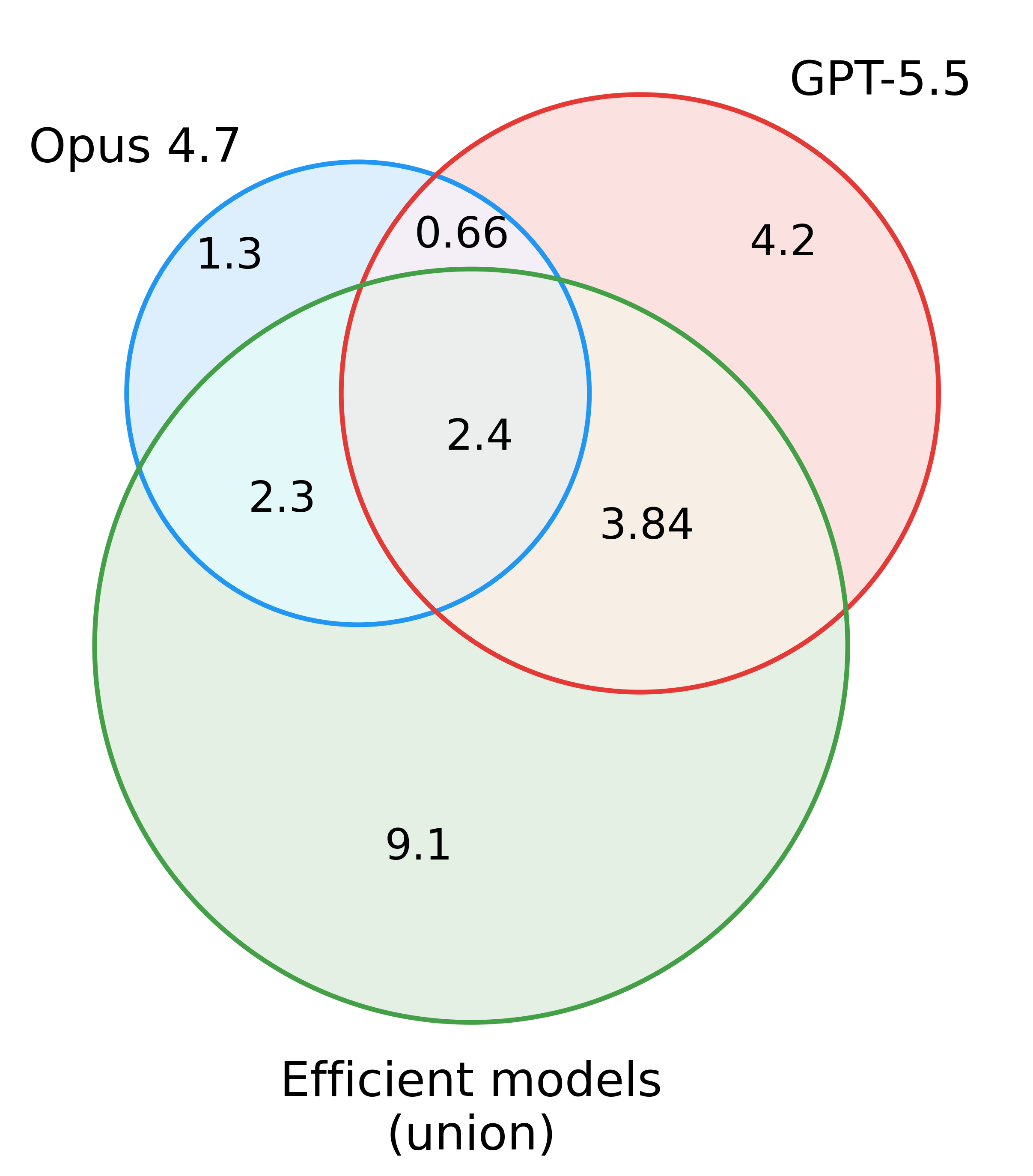}
\caption{Claude Opus 4.7, GPT-5.5, and the union of the efficient models under OpenAIReview, paragraph-level overlap on the quality-proxy papers ($70$ ICLR/NeurIPS papers from Section~\ref{sec:conference}); 3-way Jaccard $0.101$. Compare to Figure~\ref{fig:overlap_models} (main text), which shows the same three sets on the perturbation benchmark (3-way Jaccard $0.316$). Values are average unique paragraphs per paper in each region.}
\label{fig:overlap_gpt_claude_outcomes}
\end{figure}

The main text reports comment-overlap analyses on the perturbation benchmark (Section~\ref{sec:analysis}). We also ran the same analyses on the quality-proxy papers ($70$ ICLR/NeurIPS papers from Section~\ref{sec:conference}). The patterns there are different and worth recording for completeness. The two main differences are: (i)~each system contributes far more evenly to total comment volume on this corpus (overall baseline $47/33/20$ for \coarse / OpenAIReview / Reviewer3, versus $13/69/19$ on the perturbation subset), and (ii)~the three systems show clearer topic specialization (Table~\ref{tab:clusters_cp_outcomes}).
\coarse is over-represented on concrete paragraph-localized issues (notation, tables/figures, experimental setup), while OpenAIReview is concentrated on argument-level critiques (claims, formal math) and Reviewer3 on experimental and statistical-reporting issues.
The across-models comparison under OpenAIReview also flips: on the quality-proxy papers, Claude, GPT-5.5, and the union of the efficient models cover largely distinct paragraphs (3-way Jaccard $0.101$, Figure~\ref{fig:overlap_gpt_claude_outcomes}), whereas on perturbed papers they overlap heavily (3-way Jaccard $0.316$, cf.\ the main-text models Venn, Figure~\ref{fig:overlap_models}).

\begin{figure}[t]
\centering
\includegraphics[width=0.85\columnwidth]{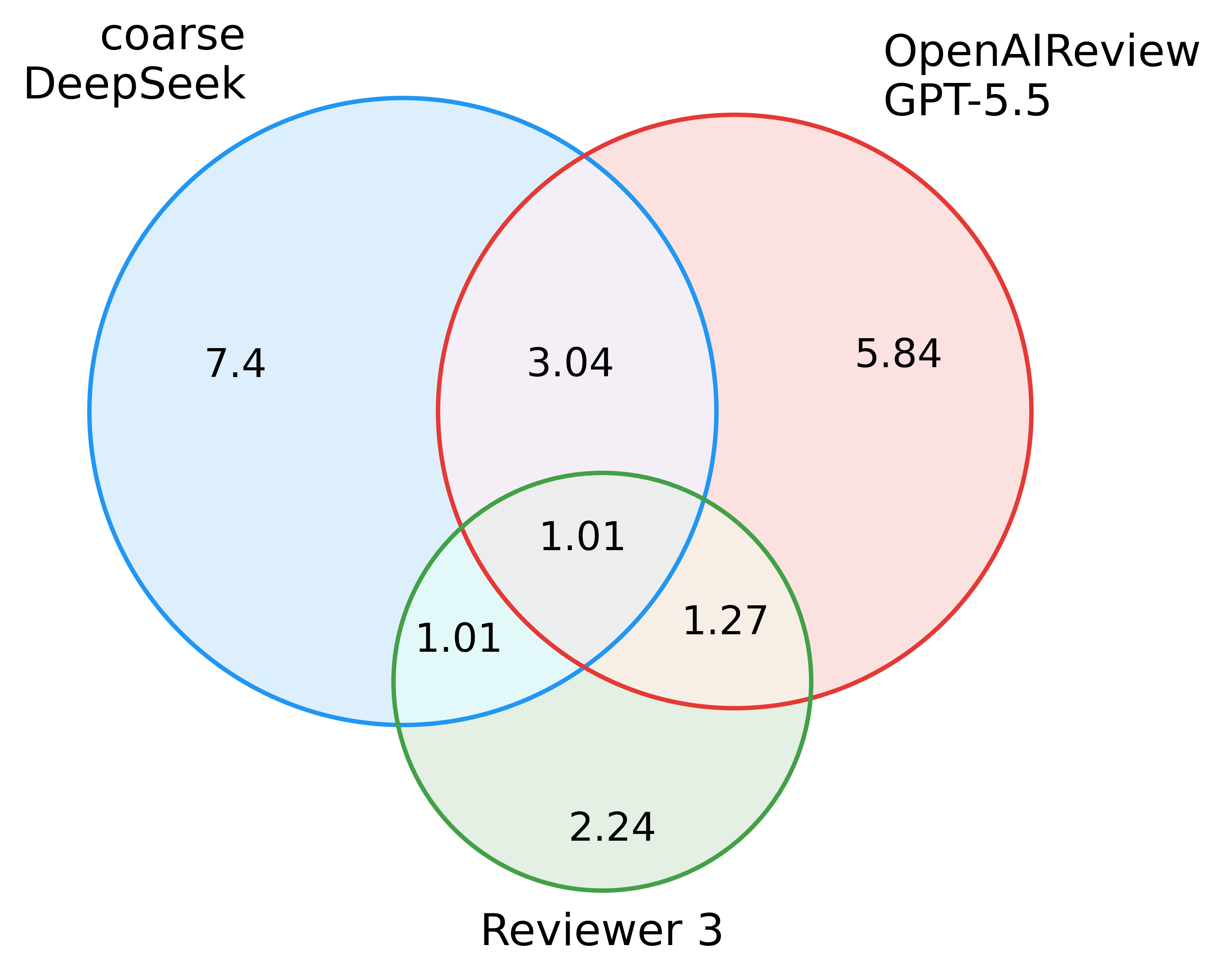}
\caption{Three-system paragraph-level overlap on the quality-proxy papers ($70$ ICLR/NeurIPS papers from Section~\ref{sec:conference}); three-way Jaccard $0.046$. Compare to Figure~\ref{fig:overlap_three_systems} (App.~\ref{app:overlap-systems}). Values are average unique paragraphs per paper in each region.}
\label{fig:overlap_three_systems_outcomes}
\end{figure}

\begin{table*}[t]
\centering
\resizebox{\textwidth}{!}{
\begin{tabular}{llccc}
\toprule
\textbf{Group} & \textbf{Representative Keywords and Phrases} & \textbf{\coarse \%} & \textbf{OpenAIReview \%} & \textbf{Reviewer3 \%} \\
\midrule
\itshape Overall baseline & \itshape (share of all comments) & \itshape 47\% & \itshape 33\% & \itshape 20\% \\
\midrule
Surface-Level (notation) & notation, equation, matrix, mathbf, dimension, undefined & 58\% & 27\% & 14\% \\
Claims \& Assertions & claim, overstates, contradiction, attention, architecture & 36\% & 39\% & 25\% \\
Experimental \& Evaluation & baseline, comparison, training, statistical, variance, GPU-hours & 46\% & 28\% & 27\% \\
Formulaic Math \& Derivations & bound, theorem, proof, convergence, manifold, equivariance & 47\% & 41\% & 13\% \\
Tables \& Figures & table, caption, figure, row, column, baseline & 62\% & 31\% & 8\% \\
\bottomrule
\end{tabular}
}
\caption{Quality-proxy version of Table~\ref{tab:clusters_cp}: \coarse is broad and concentrated on concrete paragraph-localized issues. OpenAIReview leans toward argument-level critiques (claims, formal math). Reviewer3 is narrowly focused on experimental and statistical issues. Each row sums to $\sim 100\%$ across the three systems. The baseline volumes are $47/33/20$. Cluster definitions are the same as Table~\ref{tab:clusters_cp}. Only the underlying corpus differs (quality-proxy papers from Section~\ref{sec:conference}, $70$ papers).}
\label{tab:clusters_cp_outcomes}
\end{table*}

\section{User Feedback: Downvote Error Analysis}
\label{app:downvote-errors}

This appendix gives the method and validation behind the downvote breakdown in Table~\ref{tab:downvote-errors} (Section~\ref{sec:production}).
We categorized all $283$ downvoted comments from the deployment with Gemini 3 Flash at temperature $0$, given the comment together with the capped paper text the reviewer model saw, so the judge has the same context the model had (Figure~\ref{fig:downvote-prompt}).
The labels capture the judge's reading of the likely problem under our taxonomy, and we never observe the user's actual reason for downvoting.
One paper is an outlier in this set: a single upload drew $41$ of the $283$ downvotes, far more than any other, because one user downvoted heavily.
That one user could skew the breakdown, so we recompute it with that paper removed.
The ranking is unchanged, and false positives remain the largest category at $33.9\%$.

\paragraph{Validation.}
We manually checked a stratified sample of $26$ of these comments, covering all six categories, by reading each comment against its paper.
We agreed with the judge on $23$.
All three disagreements sit on one boundary: whether a correct comment is trivial or substantive.
The judge filed one substantive catch as a false positive, and promoted two trivial points to substantive catches.
So the broad split is reliable: the dominant modes (false positives, nitpicks, and over-asking) are robust, but the line between a substantive catch and a minor one is noisy.
We therefore read the correct-comment share as an upper bound on how often a downvote lands on a genuinely useful comment.

\begin{figure*}[t]
\begin{lstlisting}[style=prompt]
You are auditing comments produced by an automated paper-review model. Each comment
below was DOWNVOTED by a real user. Your job is to assign each downvoted comment to
EXACTLY ONE category from this taxonomy, judging ONLY against the paper text provided
(that text is exactly what the review model itself saw; do not assume anything beyond it).

TAXONOMY (assign exactly one integer category per comment):
1. Parsing/OCR artifact -- the comment is about a garbled glyph / typesetting error from
   PDF conversion with NO content meaning (broken headers, image-insertion markers like
   img-0.jpeg, mangled line breaks). This is pure rendering noise. (NOT a transcription
   error that changes a number/name/citation -- that is category 5.)
2. False positive (model error) -- the model flagged a non-issue, or its own reasoning is
   wrong or "imprecise but actually fine." The flagged thing is NOT a real defect.
3. Trivial nitpick -- valid but minor: notation/caption consistency, checklist items,
   stylistic/formatting points.
4. Underspecified / deferred -- a complaint that something is too vague/underspecified, or
   that is actually addressed in an appendix / elsewhere the model didn't account for
   (reproducibility / completeness complaints).
5. Good catch, user disagreed -- the comment is correct and substantive: a real content
   error worth fixing (arithmetic-vs-table mismatch, wrong citation year/author, misleading
   aggregate, a real bug). A transcription/rendering error that CHANGES a number, name, or
   citation that the comment correctly flags belongs HERE.
6. Other / unclear -- does not fit the above, or genuinely ambiguous.

DECISION RULES:
(a) Pure glyph noise with no semantic content -> 1; but a rendering error that changes a
    number/name/citation that the comment correctly flags -> 5.
(b) False-positive vs nitpick vs good-catch turns on one question: is the flagged thing a
    REAL defect? No -> 2. Yes but minor -> 3. Yes and substantive -> 5.
(c) Judge ONLY against the provided paper text.

OUTPUT FORMAT -- return STRICT JSON and nothing else: a JSON array, one object per comment,
each object exactly:
  {"comment_id": "<the id string>", "category": <int 1-6>, "justification": "<one short sentence>"}
Include every comment_id given, exactly once. No markdown, no prose outside the JSON array.

<the paper text>

<the downvoted comments>
\end{lstlisting}
\caption{The prompt used to classify each downvoted comment into the error taxonomy.
Angle-bracketed fields are filled in per review: the paper text (the reviewer's capped paragraphs) and the downvoted comments to classify (one JSON object per comment).}
\label{fig:downvote-prompt}
\end{figure*}

\section{Artifacts, Licenses, and Compute}
\label{app:artifacts}

\subsection{Licenses and terms of use}
\label{app:licenses}
Table~\ref{tab:licenses} lists the license or terms of use for every model and dataset artifact we use. The reviewer backbones and the two pipeline LLMs (Gemini-3-Flash-Preview as generator/judge, Claude-Sonnet-4.6 as verifier) split into closed, API-only models (governed by their providers' commercial/API terms, with no released weights) and open-weight models under permissive licenses. The clustering encoder \texttt{all-MiniLM-L6-v2} is Apache-2.0. Our corpora derive from the SNOR dataset (CC-BY-4.0) and from arXiv \LaTeX{} sources, which carry each paper's individual arXiv license (most under arXiv's default perpetual, non-exclusive license, which permits research use but restricts redistribution by third parties).

\begin{table}[t]
\centering\small
\resizebox{\columnwidth}{!}{%
\begin{tabular}{@{}lll@{}}
\toprule
\bfseries Artifact & \bfseries Access & \bfseries License / Terms \\
\midrule
GPT-5.5 & API (closed) & OpenAI Business Terms \& Usage Policies \\
Claude Opus 4.7 / Sonnet 4.6 & API (closed) & Anthropic Commercial ToS \& Usage Policy \\
Gemini 3.1 Flash-Lite / 3 Flash & API (closed) & Gemini API \& Google APIs ToS \\
Grok-4.1-Fast & API (closed) & xAI Enterprise ToS \& AUP \\
DeepSeek-V4-Flash & open weights & MIT \\
Qwen3.6-35B-A3B & open weights & Apache-2.0 \\
all-MiniLM-L6-v2 & open weights & Apache-2.0 \\
SNOR dataset~\citep{snor} & data & CC-BY-4.0 \\
arXiv paper sources & data & per-paper arXiv license \\
\bottomrule
\end{tabular}}
\caption{Licenses / terms of use for the models and datasets used. Closed models are accessed only through their providers' APIs under the listed commercial terms; open-weight models and datasets are under the listed licenses.}
\label{tab:licenses}
\end{table}

\subsection{Consistency with intended use}
\label{app:intended-use}
Our use of every artifact is consistent with its intended use. The closed models are accessed through their official APIs for research evaluation, which the providers' commercial/API terms permit; the open-weight models and \texttt{all-MiniLM-L6-v2} are used under permissive licenses (MIT, Apache-2.0) that allow research use; and SNOR is used under CC-BY-4.0 with attribution. The perturbation corpus is built from publicly available arXiv \LaTeX{} sources accessed for research. The artifact we create (the perturbation benchmark) is intended for research use only. Because it derives from data accessed for research (arXiv sources and OpenReview-derived signals via SNOR), we restrict it to research contexts and, to respect the default arXiv license, intend to release the injected edits and metadata keyed by arXiv identifier rather than redistributing full paper text.

\subsection{Model sizes, infrastructure, and cost}
\label{app:compute}
The closed API models (GPT-5.5, Claude Opus 4.7 / Sonnet 4.6, Gemini 3.1 Flash-Lite / 3 Flash, Grok-4.1-Fast) do not have publicly disclosed parameter counts. Among the open-weight models, Qwen3.6-35B-A3B is a mixture-of-experts model with $35$B total ($\sim$$3$B active) parameters, DeepSeek-V4-Flash is an open-weight mixture-of-experts model (parameter count per its model card), and \texttt{all-MiniLM-L6-v2} has $22.7$M parameters. We did not train or fine-tune any model: all reviewer systems were run through hosted inference APIs (OpenRouter and the providers' own endpoints), so our compute budget is API inference rather than local GPU-hours. The only local computation is sentence embedding and $k$-means clustering for the comment analysis, which runs in minutes on a single machine.

Table~\ref{tab:cost} reports the API cost of the \emph{evaluation} (review) runs, totalling roughly \$2{,}500 and $353$M tokens. Benchmark construction (perturbation generation, verification, and scoring) and the comment-clustering analysis incur additional, smaller API costs not included here.

\begin{table}[t]
\centering\small
\begin{tabular}{@{}lrr@{}}
\toprule
\bfseries Evaluation runs & \bfseries Cost & \bfseries Tokens \\
\midrule
Quality-proxy, frontier subset    & \$204 & 21M \\
Quality-proxy, efficient models    & \$256 & 173M \\
Perturbation benchmark (all systems)  & \$2{,}049 & 159M \\
\midrule
\bfseries Total & \bfseries \$2{,}509 & \bfseries 353M \\
\bottomrule
\end{tabular}
\caption{Approximate API cost and token usage of the evaluation (review) runs, from logged per-call costs (OpenRouter / provider billing). Excludes benchmark-construction and clustering costs.}
\label{tab:cost}
\end{table}